\documentclass[times,review,10pt]{elsarticle}
\biboptions{sort&compress}
\usepackage{xspace}
\usepackage{acronym}
\usepackage{amssymb}
\usepackage{amsmath}
\usepackage[hidelinks]{hyperref}
\usepackage[capitalise]{cleveref}
\usepackage{booktabs}
\usepackage{multirow}
\usepackage[dvipsnames]{xcolor}
\usepackage{caption}
\usepackage{lineno}
\usepackage{listings}
\usepackage{pifont}
\usepackage{makecell}
\usepackage{marvosym}
\usepackage[table]{xcolor}
\usepackage{threeparttable}

\graphicspath{{figures/}{pictures/}{images/}{./}}

\makeatletter
\DeclareRobustCommand\onedot{\futurelet\@let@token\@onedot}
\def\@onedot{\ifx\@let@token.\else.\null\fi\xspace}

\makeatother

\crefname{algorithm}{Alg.}{Algs.}
\Crefname{algocf}{Algorithm}{Algorithms}
\crefname{section}{Sec.}{Secs.}
\Crefname{section}{Section}{Sections}
\crefname{table}{Tab.}{Tabs.}
\Crefname{table}{Table}{Tables}
\crefname{figure}{Fig.}{Fig.}
\Crefname{figure}{Figure}{Figure}

\journal{Pattern Recognition}

\begin{document}

\begin{frontmatter}



\title{SFGFusion: Surface Fitting Guided 3D Object Detection with 4D Radar and Camera Fusion}

\cortext[cor]{Corresponding author}
\author[label1,label3]{Xiaozhi Li}
\author[label2]{Huijun Di\corref{cor}}
\author[label1,label4]{Jian Li}
\author[label5]{Feng Liu}
\author[label2]{Wei Liang}

\affiliation[label1]{organization={Radar Technology Research Institute, School of Information and Electronics, Beijing Institute of Technology},
            postcode={100081},
            city={Beijing},
            country={China}}

\affiliation[label2]{organization={School of Computer Science and Technology, Beijing Institute of Technology},
            postcode={100081},
            city={Beijing},
            country={China}}
            
\affiliation[label3]{organization={Innovative Equipment Research Institute, Beijing Institute of Technology},
            postcode={610299},
            city={Chengdu},
            country={China}}

\affiliation[label4]{organization={Key Laboratory of Electronic and Information Technology in Satellite Navigation (Beijing Institute of Technology), Ministry of Education},
            postcode={100081},
            city={Beijing},
            country={China}}

\affiliation[label5]{organization={Beijing Racobit Electronic Information Technology Co., Ltd.},
            postcode={100097},
            city={Beijing},
            country={China}}

\begin{abstract}
3D object detection is essential for autonomous driving. As an emerging sensor, 4D imaging radar offers advantages as low cost, long-range detection, and accurate velocity measurement, making it highly suitable for object detection. However, its sparse point clouds and low resolution limit object geometric representation and hinder multi-modal fusion. In this study, we introduce SFGFusion, a novel camera-4D imaging radar detection network guided by surface fitting. By estimating quadratic surface parameters of objects from image and radar data, the explicit surface fitting model enhances spatial representation and cross-modal interaction, enabling more reliable prediction of fine-grained dense depth. The predicted depth serves two purposes: 1) in an image branch to guide the transformation of image features from perspective view (PV) to a unified bird’s-eye view (BEV) for multi-modal fusion, improving spatial mapping accuracy; and 2) in a surface pseudo-point branch to generate dense pseudo-point cloud, mitigating the radar point sparsity. The original radar point cloud is also encoded in a separate radar branch. These two point cloud branches adopt a pillar-based method and subsequently transform the features into the BEV space. Finally, a standard 2D backbone and detection head are used to predict object labels and bounding boxes from BEV features. Experimental results show that SFGFusion effectively fuses camera and 4D radar features, achieving superior performance on the TJ4DRadSet and view-of-delft (VoD) object detection benchmarks.
\end{abstract}


\end{frontmatter}


\section{Introduction}\label{sec:introduction}
\sloppy
In recent years, the rise of autonomous driving and deep learning technology has led to a rapid development of 3D object detection. As an essential component of autonomous driving, 3D object detection is required to provide accurate and reliable perception information for downstream tasks such as prediction, decision-making, planning and control \cite{survey_autonomous_driving}.

\begin{figure}[t]
\centering{\includegraphics[scale=0.26]{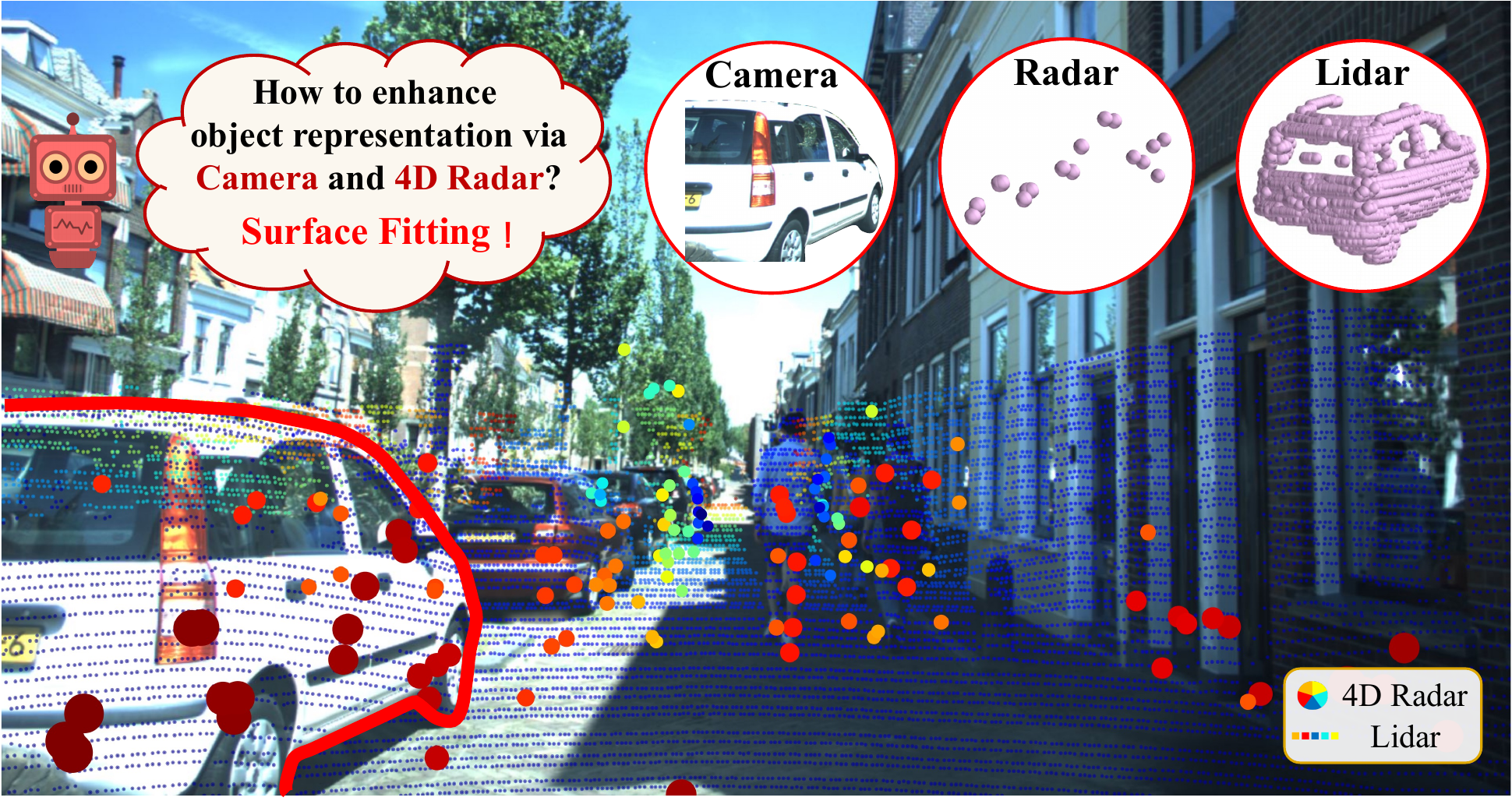}}
\caption{\textbf{Projection of 4D radar and LiDAR point clouds onto image.} The three modalities of the same object are highlighted. The radar point cloud is sparse and irregular, while the LiDAR point cloud is denser and structured.}
\label{Fig_4D_Radar_LiDAR_Contrast}
\end{figure}

To enhance the environmental perception capabilities of object detection algorithms, multi-modal fusion techniques are increasingly employed. Common sensors include cameras, LiDAR, and millimeter-wave radar, each offering distinct data structures and attributes. As an emerging technology, 4D imaging radar adds elevation information and improves resolution, significantly increasing point cloud density\cite{survey_4d_radar}. Although 4D imaging radar still trails behind LiDAR in terms of point cloud density and quality (as shown in   \cref{Fig_4D_Radar_LiDAR_Contrast}), its advantages in speed measurement, cost efficiency, and all-weather operation position it favorably for applications in object detection. Furthermore, integrating camera and 4D imaging radar features to leverage the advantages of multi-modal data can enhance object representation, providing comprehensive and reliable environmental information for autonomous driving systems \cite{RCFusion,LXL,RCBevdet}.

In multi-modal 3D detection frameworks, accurate alignment of features from different modalities is critical, especially the process of image feature transformation from 2D planes to 3D space \cite{survey_3D_detection}. Since image data lacks depth information, accurate depth estimation during feature view transformation is crucial, as it directly influences the precision of spatial feature mapping. To improve depth estimation, point cloud data can serve as a guiding reference during the transformation process \cite{bevfusion,focal}. This strategy is effective for LiDAR, as its point clouds are dense and systematically distributed across the scene. In contrast, point clouds from 4D imaging radar are typically sparse and irregular, making it difficult to provide sufficient geometric constraints in object contours and key regions. This results in weaker guidance during the view transformation of image features, presenting challenges to the precise alignment of features between the camera and radar.

Furthermore, during point cloud feature extraction, the sparsity limitations of 4D imaging radar hinder its ability to accurately depict object surface shapes and contour features. This leads to challenges for the backbone \cite{pointnet,pointpillars,Voxelnet} in extracting sufficient informative features from the sparse point clouds, which fail to meet the requirements of object detection algorithms. Moreover, the irregularity and sparsity of the point clouds result in radar features that are often deficient in density and structural integrity, thereby limiting their capacity to provide accurate geometric representations. This impairs the fusion of radar and image features and may even reduce the quality of the image features, ultimately impacting the overall algorithm performance.

Addressing the challenges discussed above, this paper proposes SFGFusion, a novel 3D object detection network that integrates camera and 4D imaging radar data, guided by surface fitting. An explicit surface fitting model is leveraged in the network to constrain and regularize the depth prediction for each foreground object. As a result, the network can more reliably predict fine-grained dense depth from the 2D image and the sparse 3D radar points. This depth information drives two key processes: 1) in an image branch to guide the transformation of image features from perspective view (PV) to a unified bird’s-eye view (BEV) perspective for multi-modal fusion, improving the precision of spatial feature mapping, and 2) in a surface pseudo-point branch to generate a dense pseudo-point cloud from the predicted depths, mitigating the limitations of the 4D imaging radar’s sparse point cloud. Meanwhile, the original 4D radar point cloud is encoded in a separate radar branch. Both point cloud branches utilize a pillar-based method and subsequently transform the corresponding features into the BEV space. Finally, a standard 2D backbone and detection head are employed to predict object labels and bounding boxes based on the BEV features.

The proposed SFGFusion was validated on the TJ4DRadSet \cite{tj4dradset} and view-of-delft (VoD) \cite{vod} datasets, demonstrating a notable performance advantage over existing algorithms and confirming its effectiveness. In addition, comparisons of different surface fitting methods and subsequent applications are made through ablation studies, demonstrating the superiority and robustness of the proposed components.

\section{Related Work} \label{sec:related work}

\subsection{Image based 3D Detection}

 Monocular images provide abundant semantic and color information, yet the lack of depth information poses a significant challenge for 3D object detection. Existing image-based methods can be divided into three categories. The first employs end-to-end networks that directly extract image features and predict object classes and positions from key points \cite{imvoxelnet,adaptive}. The second projects image features from the PV to the BEV perspective, then applies 2D feature extraction modules and detection heads to yield detection results from a top-down perspective \cite{bevformer,bevformerv2}. The third estimates pixel-level depth to project 2D images into 3D space, generating pseudo-point clouds that can be processed with conventional detection heads \cite{pseudo-lidar,pseudo-mono}.

\subsection{Point Cloud based 3D Detection}

LiDAR, with its high precision and resolution, is the primary sensor for point cloud–based 3D detection. Owing to the unordered and dense nature of LiDAR point cloud, it typically requires encoding into specific formats for feature extraction. Pillar-based methods \cite{pointpillars,pillarnext} discretize point cloud data into fixed-size 2D grids, preserving geometric and clustering cues while reducing dimensionality. Voxel-based methods \cite{lgnet,hrnet} convert point cloud data into regular 3D grids, retaining richer height information and offering better representation for dense point cloud. Both methods reduce computational costs and memory usage but introduce quantization errors and information loss. Point-based methods \cite{pointnet,smurf} directly process the raw points, which retain detailed features but encounter high computational and memory costs. After encoding, the detection head is used to determine the object class and position, with common types including anchor-based \cite{pointpillars,Voxelnet,second} and anchor-free \cite{centerpoint,3d-centernet,spatial}. While these components are generally applicable to point clouds from various sensors, selecting the appropriate components based on the characteristics of point cloud is crucial in practical applications.

Compared to LiDAR, which provides point cloud positional information, ordinary automotive radar also offers reflectivity and velocity data. However, the point cloud from the ordinary automotive radar is sparse and lacks elevation information, making 3D object detection tasks challenging in the radar-only modality \cite{radardistill}. In contrast, 4D imaging radar significantly improves point cloud density and includes elevation information, greatly enhancing the potential for radar-based detection tasks. RPFA-Net~\cite{RPFA-Net} takes advantage of the multi-dimensional information in the 4D radar point cloud, dividing the point cloud into pillars and incorporating an attention mechanism during encoding, effectively improving detection performance. RadarMFNet \cite{RadarMFNet} employs multi-frame point clouds for detection, leveraging both temporal and spatial features to improve detection accuracy. SMURF \cite{smurf} introduces kernel density estimation to reduce the impact of point cloud sparsity, enhancing the accuracy of 4D radar-only object detection.

\subsection{Fusion based 3D Detection} 

Recent studies widely adopt multi-modal fusion to combine image semantics with point cloud geometry, boosting 3D detection performance. Typical approaches include point feature expansion \cite{pointaugmenting,multimodal}, region of interest (RoI) feature fusion \cite{mv3d,mvx-net}, BEV feature fusion \cite{RCBevdet,bevfusion}, and virtual point feature fusion \cite{SFD,virtual-sparse}. The point feature expansion method involves extending point cloud features to image, which retains the fine-grained features of the point cloud while adding corresponding image texture information. However, this method requires point-wise computation, leading to high computational costs. RoI feature fusion integrates features at the RoI level, effectively combining local information from images and point clouds. It offers flexibility in choosing fusion strategies, but the detection accuracy is highly influenced by the quality of RoI generation. BEV feature fusion merges modalities in the bird's eye view perspective, efficiently capturing global information and spatial relationships. The resulting fused features can be directly processed by mature backbone and detection heads, though this method necessitates a reasonable and precise approach for feature view transformation. Virtual point feature fusion enriches the original point cloud by generating pseudo-points from images, thereby enhancing object representation. The effectiveness of this fusion depends on the quality of pseudo-point generation and requires guidance from the original dense point cloud.

Current research has made significant strides in multi-modal fusion and feature extraction methods. However, the majority focuses on camera and LiDAR, with fewer works on camera and 4D imaging radar. Due to significant differences in the generation principles and distribution characteristics between the radar and LiDAR point clouds, directly applying LiDAR-based modules often yields suboptimal detection. Addressing the sparsity and unordered nature of the radar point cloud necessitates a redesign of various detector modules. RCFusion \cite{RCFusion} fuses image and radar point cloud features in BEV, projecting image features into 3D space via an orthographic transform and extracting radar features based on position, velocity, and reflectivity. LXL \cite{LXL} uses a sampling method for image feature conversion, aided by a radar occupancy grid for improved view transformation and multi-modal interaction. RCBEVDet \cite{RCBevdet} employs point-based and transformer-based encoders for enhanced feature extraction, combined with a cross-attention fusion module to facilitate interaction across modalities. Parallel to the above camera-radar fusion methods, this paper focuses on fitting object surface parameters to improve the precision of the image feature view transformation and mitigate the limitation of the sparse radar point cloud.

\section{Method} \label{sec:method}

\begin{figure*}[t!]
\centering{\includegraphics[width=\textwidth, height=\textheight, keepaspectratio]{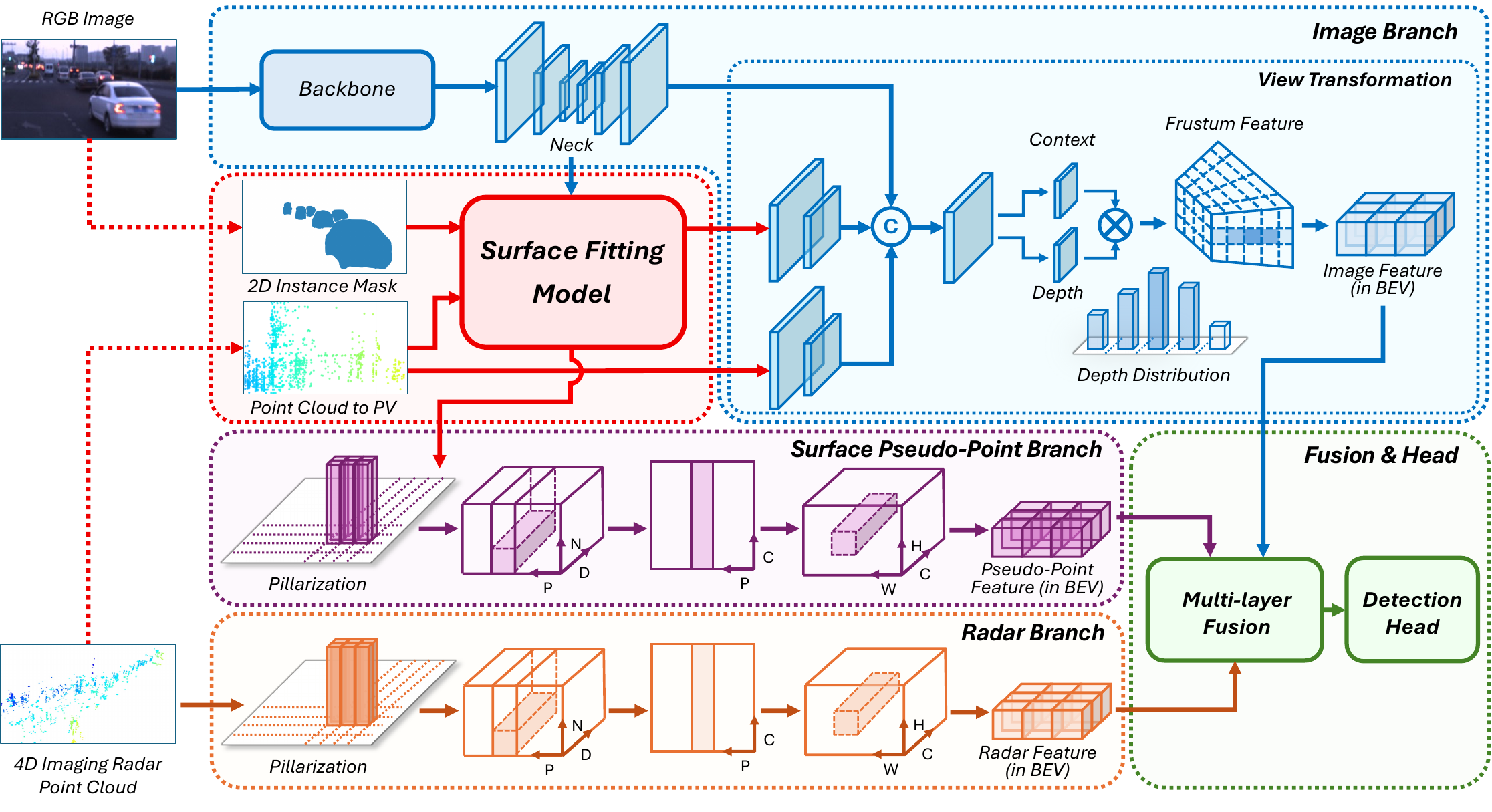}}
\caption{\textbf{The overall architecture of the SFGFusion.} In the surface fitting model, we integrate image semantic information with point cloud spatial data to predict the fine-grained dense depth of object surfaces. In the image branch, the surface fitting dense depth and the 4D radar point cloud guide the view transformation of the image feature, enhancing depth estimation accuracy. The surface pseudo-point branch converts the depth data from surface fitting into dense pseudo-point cloud and performs feature extraction to compensate for the sparse nature of 4D radar point cloud. In the radar branch, both point cloud spatial data and radar-specific velocity and RCS features are exploited to increase feature diversity. Finally, the fused BEV features from all branches are processed through a multi-scale network, and the detection head is employed to generate 3D bounding boxes, object classifications, and confidence scores.
}
\label{Fig: SFGFusion model architecture}
\end{figure*}

We propose SFGFusion, a 3D object detection network designed for the fusion of image and 4D imaging radar features. Its overall architecture is illustrated in \cref{Fig: SFGFusion model architecture}. The proposed framework comprises five parts: a surface fitting model that estimates the dense depth information of objects, an image branch with view transformation guided by surface fitting depth information, a radar branch with multi-dimensional point cloud information processing, a surface pseudo-point branch with instance-level dense 3D spatial information processing, and a fusion module as well as detection head. Due to the distinct coordinate systems of images and radar point clouds, direct spatial alignment of multi-modal information is infeasible. Thus, we perform feature interaction and fusion in the BEV domain.

\subsection{Surface Fitting Model}
\begin{figure*}[t]
\centering{\includegraphics[width=\textwidth, height=\textheight, keepaspectratio]{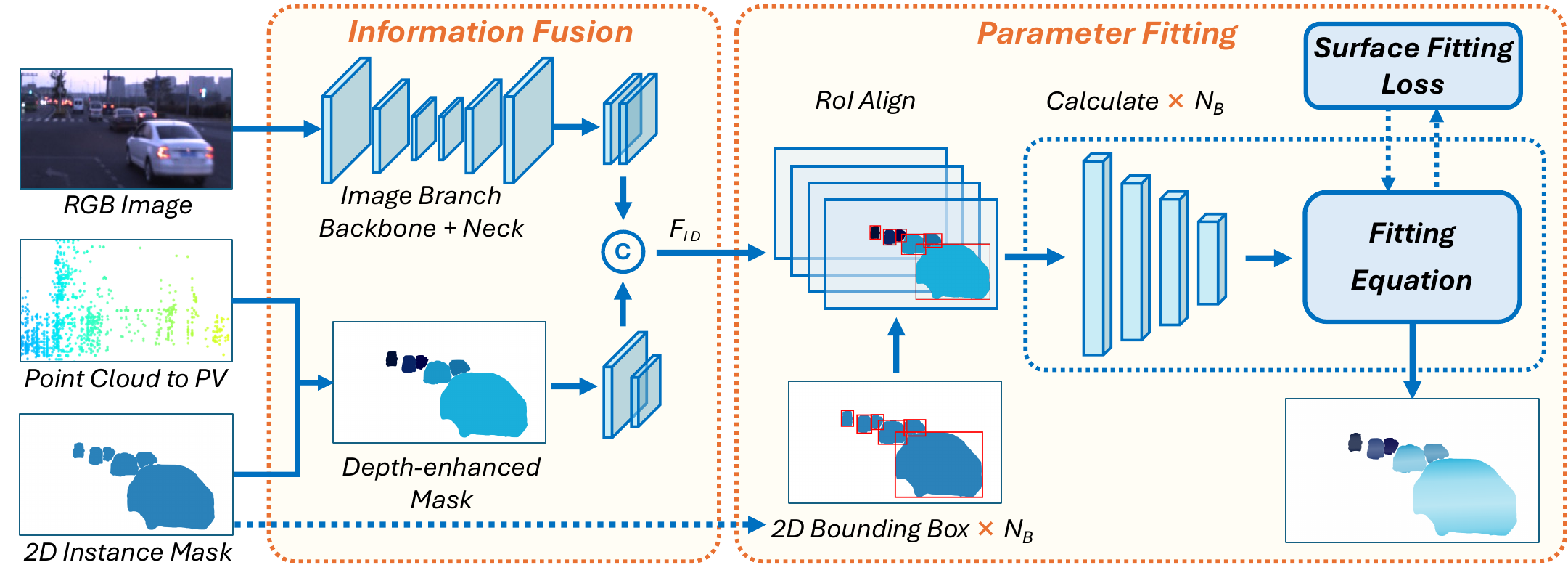}}
\caption{\textbf{The surface fitting model in SFGFusion.} During information fusion, the depth-enhanced mask is generated by averaging the radar point depths within each 2D instance mask and then concatenated with the original image feature from the image branch, facilitating interaction between the image and radar modalities. For parameter fitting, the 2D bounding box generated from the instance mask is used to guide a lightweight MLP network in estimating the coefficients of the surface fitting equation, which is then used to calculate the depth of each pixel within the instance. The depth estimation is refined by supervising the surface fitting using the radar point cloud data, leading to fine-grained and accurate depth prediction for each instance.
}
\label{Fig: Surface Fitting model architecture}
\end{figure*}

The surface fitting model aims to estimate the object's surface depth by integrating image semantics and point cloud spatial information. As illustrated in \cref{Fig: Surface Fitting model architecture}, the model consists of two components: information fusion and parameter fitting. The information fusion module combines image, point cloud, and instance mask data from the perspective view, providing a robust feature representation for parameter fitting. The parameter fitting module extracts foreground object features within category-agnostic 2D bounding boxes using ROI Align and employs a lightweight network to estimate fitting equation parameters. Leveraging the surface fitting equation, the model estimates pixel-level depth values, enhancing image feature projection accuracy and compensating for point cloud sparsity, ultimately improving multi-modal fusion performance.

\subsubsection{Information Fusion} 
To effectively integrate the semantic information of the image and the geometric information of the radar point cloud, we preprocess the input data before surface parameter fitting. Formally, given a point cloud and an image as input, we first project the 3D points onto 2D image and preserve points falling within the 2D foreground instance masks. These 3D points are kept as reference points to provide depth and are represented as $R = \{ (u_i, v_i, d_i) \}_{i=1}^{N_R}$, where $u_i$ and $v_i$ are pixel coordinates, and $d_i$ is the actual depth. Owing to radar point cloud sparsity, each foreground instance mask contains only a few reference points. To supplement the depth information, we assign the average depth of reference points as the depth value for the entire mask. This value is used to fill in all pixel values within the mask, generating a depth-enhanced mask image $M_D$, thereby incorporating depth information into the image features.

While the above method is effective, the sparse 4D radar point cloud data for each instance, combined with the calibration accuracy of the radar-camera transfer matrix, adversely affects the accuracy of pixel depth estimation within each instance. Therefore, in addition to using the average depth of the point cloud for depth supplementation, we incorporate the texture and semantic information from the original image. The image feature $F_I$ extracted by the backbone and neck models in the image branch is further processed to create a feature map $F_{IS}$ suitable for surface fitting. The feature map $F_{IS}$ is then concatenated with the resized depth-enhanced mask image $M_D$, generating a fusion feature map $F_{ID}$ with both depth and texture information. This process is described as:
\begin{equation}
F_{ID} = Concat(CBR(F_I),CBR(M_D)),
\end{equation}
where $CBR(\cdot)$ denotes the sequential application of Convolution, Batch Normalization, and ReLU activation, $Concat(\cdot)$ concatenates the feature along the channel dimension.

To draw more attention to the foreground instances, we extract the 2D bounding boxes without category information $B = \{(u_{j1}, v_{j1}, u_{j2}, v_{j2}) \}_{j=1}^{N_B}$  based on their minimum and maximum pixel coordinates. These bounding boxes are used for feature localization in the parameter fitting process. 

\subsubsection{Parameter Fitting}
After information fusion, the surface fitting model takes the fusion feature map $F_{ID}$ as input and uses 2D bounding boxes as guidance. The RoI Align module \cite{mask-rcnn} identifies the features corresponding to each instance, and a lightweight MLP network combined with a fitting equation calculates the surface depth of the instances. 

In the surface parameter fitting process for one instance, the RoI Align module extracts multi-channel feature data from the bounding box in the feature map $F_{ID}$ and resizes it to a uniform size. This 2D feature is then flattened into a 1D vector, and the MLP transforms it to obtain the fitting equation coefficients $\rho$. The above process can be mathematically described as follows:
\begin{equation}
\rho = MLP(Flatten(RoI\_Align[F_{ID},B])),
\end{equation}
where $RoI\_Align[\cdot]$ extracts the feature within the bounding box, $Flatten(\cdot)$ flattens the 2D feature into 1D vector, and $MLP(\cdot)$ represents the Multi-Layer Perceptron network.

The fitting equation takes all pixel coordinates ${(u_s,v_s)}$ in the instance mask as input and outputs the depth values $d_s$. To effectively model the underlying geometry of each object instance, we adopt a quadratic surface equation in 3D space as the fitting equation. Compared to plane or linear models, the quadratic surface offers greater geometric modeling capacity, allowing for the representation of curved object surfaces commonly found in real-world scenes (e.g., vehicles, pedestrians). The specific form of the fitting equation is:
\begin{equation}
\label{equ_surf}
d_s = a_{c} \cdot u_s^{2} + b_{c} \cdot v_s^{2} + c_{c} \cdot u_s \cdot v_s + d_{c} \cdot u_s + e_{c} \cdot v_s + f_{c},
\end{equation}
where $\rho = [a_c,b_c,c_c,d_c,e_c,f_c]$ is the coefficients of the fitting equation. By fitting a smooth quadratic surface guided by both image semantics and radar geometry, the model can generate more accurate and continuous pixel-level depth estimations.

To improve the precision of surface fitting, the supervised training process must be carefully managed, particularly the design of the loss function. In training, the sparse point cloud depth values and the average point cloud depth are used as ground truth for supervision, and the fitting equation calculates the predicted depths for all pixel positions within the mask. For each frame, the loss values of all instances are computed and summed to produce the surface fitting loss $L_{sf}$. The loss formula is:
\begin{equation}
L_{sf}=\sum_{j=1}^{N_{B}}\left(\frac{1}{N_{R}}\sum_{i=1}^{N_{R}}(d_{i,j}^{t}-d_{i,j}^{p})^2+\lambda(\overline{d_{j}^{t}}-\overline{d_{j}^{p}})^2\right),
\end{equation}
where $N_B$ is the number of bonding boxes in the current frame,  $N_R$  is the number of reference points $R$ in each instance, $d_{i,j}^{t}$ is the depth value of the reference points $R$, $d_{i,j}^{p}$ is the predicted depth value by the fitting equation at the corresponding image location of the reference points, $\overline{d_{j}^{t}}$ is the average depth of the reference points $R$, and $\overline{d_{j}^{p}}$ is the average predicted depth within the whole instance mask.

Once the parameters of the fitting equation are determined, \cref{equ_surf} is used to predict the depth $d_{s}$ for all pixel positions within the corresponding mask, completing the depth augmentation for each instance mask and generating fine-grained instance-level depth estimation features. These depth features will guide the image view transformation and 3D spatial feature extraction in the surface pseudo-point branch.

\subsection{Image Branch}
The image branch comprises three components: backbone, neck, and view transformation. Initially, the backbone processes the 2D image to extract low-level features. These features are combined across multiple scales through the neck, generating a multi-level image feature $F_I$ for view transformation and surface fitting. The view transformation, guided by the surface fitting depth information and the 4D radar point cloud, projects the image feature into 3D space and compresses it along the height dimension to generate the BEV feature $F_{BEV}^I$ of the image branch.
\subsubsection{Backbone and Neck}
We selected Swin-T \cite{Swin-T} as the backbone for image feature encoding. Next, the image features are aggregated in the neck using FPN \cite{FPN}, which combines low-level detailed features with high-level global features while reducing the feature map to 1/8 of its original size.

\subsubsection{View Transformation} \label{Section-View Transformation} 
To achieve a precise transformation of image feature from the PV to BEV perspective, we have enhanced the view transformation method \cite{lss} by integrating the image feature, depth feature from the surface fitting model, and radar feature from the 4D radar point cloud projected onto the PV perspective. Initially, the depth feature $F_{SF}$ and radar feature $F_{PR}$ are processed through feature extraction to standardize their dimensions with the image feature $F_I$. Once aligned, the three feature maps are concatenated to construct the combined feature $F_v$, formulated as follows:
\begin{equation}
F_{v} = Concat(F_I,CBR(F_{SF}),CBR(F_{PR})),
\end{equation}
where $F_v$ of size $(C_v,H_v,W_v)$ serves as the input for subsequent steps.

\begin{figure}[t!]
\centering{\includegraphics[scale=0.25]{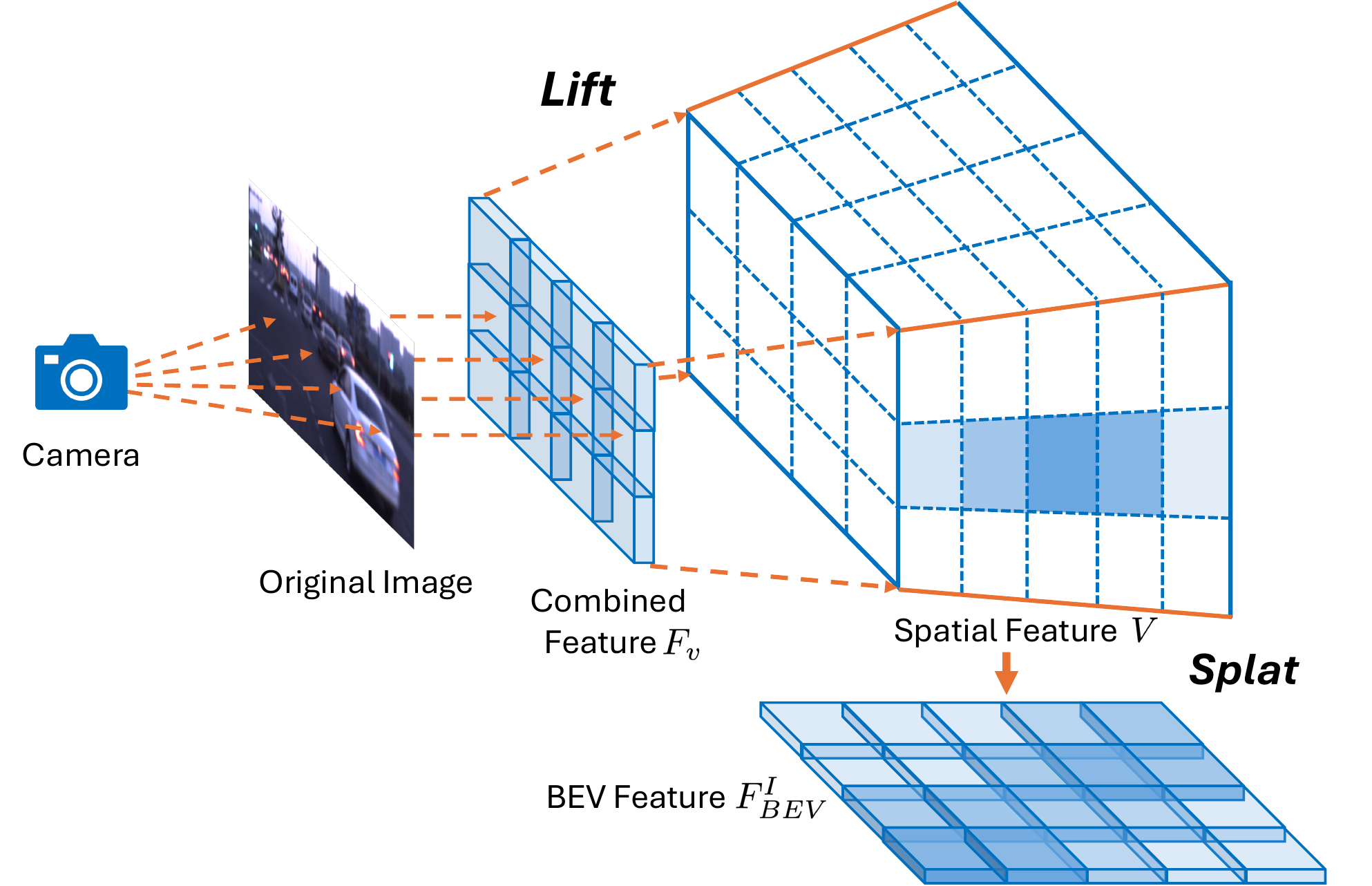}}
\caption{\textbf{Illustration of image feature view transformation.} The process comprises two steps: Lift and Splat. The combined feature $F_v$ integrates information from the image, point cloud, and surface fitting model.}
\label{Fig: View Transformation}
\end{figure}

As shown in \cref{Fig: View Transformation}, the view transformation consists of two steps. “Lift” projects the PV-view combined feature $F_v$ into the 3D frustum space based on depth distribution probability. “Splat” applies BEV pooling to compress the 3D feature along the height dimension, generating the BEV feature $F_{BEV}^I$ of the image branch.

\textbf{Lift:} Given the combined feature $F_v$, \cref{equ_depth} is used to estimate the discrete depth distribution $P_D$ along the camera ray for each feature pixel at resolution $(H_v, W_v)$:
\begin{equation}
\label{equ_depth}
P_D = \psi(F_v),
\end{equation}
where $\psi$ represents a lightweight DepthNet. The size of $P_D$ is $(D_b,H_v, W_v)$, with $D_b$ denoting the number of depth bins.

Next, the context information of the combined feature is associated with the depth distribution to obtain the frustum feature $F_u$ in 3D space, which can be described as: 
\begin{equation}
F_u = F_v \otimes P_D,
\end{equation}
where $\otimes$ denotes the outer product, and the size of $F_u$ is $(C_v,D_b,H_v, W_v)$.

Finally, the frustum feature $F_u$ is projected into the 3D voxel space using the camera’s intrinsic and extrinsic matrices, resulting in the spatial feature $V$ with dimensions $(C_v, X_v, Y_v, Z_v)$.

\textbf{Splat:} BEV Pooling \cite{bevdet} is applied to compress the spatial feature $V$ from the 3D voxel space into a single height plane, generating a pseudo-image of shape $(C_I, H, W)$, which serves as the BEV feature $F_{BEV}^I$ for the image branch.

\subsection{Radar Branch}
The radar branch takes the 4D radar point cloud as input. Due to the imaging principles and resolution of the sensor, the 4D point cloud density is sparser than LiDAR. Using a 3D voxel processing method akin to VoxelNet\cite{Voxelnet} would result in a large number of empty voxels. Moreover, the 4D imaging radar has a low resolution in terms of height dimension. Given this sparsity, the pillar-based partitioning method is more effective for feature aggregation in the height direction. In this section, we utilize PointPillars \cite{pointpillars} as the backbone for feature extraction. The input 3D point cloud is initially discretized in the X-Y plane, dividing it into pillars based on their 2D projection coordinates. Subsequently, the features of each point within a pillar are enhanced by incorporating the unique intensity and velocity information of the 4D radar point cloud. Each point $P_r$ in the radar point cloud has the following attributes:
\begin{equation}
\label{deqn_ex1}
P_r = 
\begin{cases} 
[x,y,z,s,v,r,\mu,\tau,x_c,y_c,z_c,x_p,y_p] & \text{TJ4D} \\
[x,y,z,s,v,v_{comp},x_c,y_c,z_c,x_p,y_p] & \text{VoD}
\end{cases}
,
\end{equation}
where [\textit{x}, \textit{y}, \textit{z}] are the spatial coordinates of the radar point, \textit{s} is the intensity information (radar cross-section or signal-to-noise ratio), \textit{v} is the relative radial velocity, $v_{comp}$ is the object’s absolute radial velocity after vehicle speed compensation, \textit{r} is the detection range to the radar center, $\mu$ and $\tau$ are horizontal angle and vertical angle of the detection. The subscript \textit{c} denotes the arithmetic mean distance from all points in the pillar, and the subscript \textit{p} indicates the deviation from the pillar center. Following this, the point cloud features are expanded to dimension $D_r$.

After expanding the feature dimensions, the number of points per pillar $N_r$ is set, using random sampling or zero-padding to ensure each of the non-empty pillars $P_p$ contains $N_r$ points, resulting in a dense tensor $F_r$ of size $(D_r, P_p, N_r)$. These features are further processed using a network consisting of linear layers, normalization, and ReLU activation functions, transforming the feature size to $(C_r, P_p, N_r)$. Applying max-pooling along the $N_r$ dimension reduces the feature size to $(C_r, P_p)$. Finally, the pillar features are projected back to their original locations, creating a pseudo-image of size $(C_r, H, W)$, which serves as the BEV feature $F_{BEV}^{R}$ of the radar branch. The above process can be represented as follows:
\begin{equation}
F_{BEV}^{R} = Proj(MaxPool(LBR(F_r))),
\end{equation}
where $LBR(\cdot)$ denotes the sequential application of linear layers (Linera), Batch Normalization (BN), and ReLU activation, $MaxPool(\cdot)$ is the max-pooling operation, and $Proj(\cdot)$ represents the pillar feature projection process.
\subsection{Surface Pseudo-Point Branch}

\begin{figure}[t]
\centering{\includegraphics[scale=0.27]{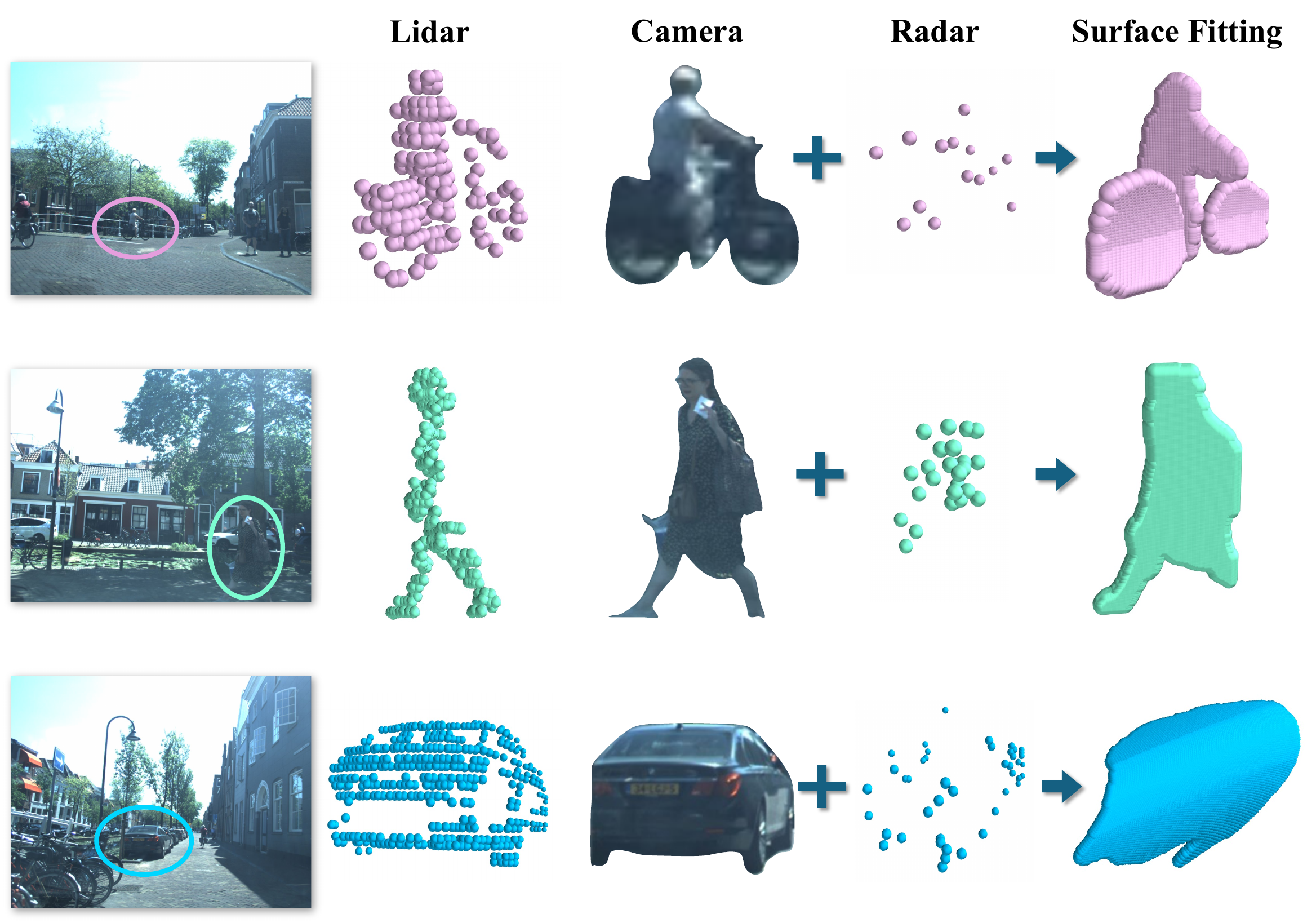}}
\caption{\textbf{Surface fitting pseudo-point visualization.} The surface fitting model generates denser 3D spatial information than LiDAR by fusing camera and radar data. Each row shows the RGB image, LiDAR point cloud, camera instance mask, 4D radar point cloud, and surface fitting pseudo-point cloud.}
\label{Fig: Surface Fitting Vis}
\end{figure}

The point cloud density of 4D imaging radar is considerably lower than that of LiDAR, making it challenging for the radar branch to extract features related to object contours and categories effectively. In the surface fitting model, we enhance instance-level depth features using both image and 4D radar point cloud, acquiring the depth $d_{s}$ for all pixel positions within the 2D foreground instance masks. To fully exploit these depth features, we project the pixel depths from the instance masks into 3D space via the radar-camera transfer matrix, generating the dense surface fitting pseudo-point cloud, as illustrated in the \cref{Fig: Surface Fitting Vis}. This pseudo-point cloud is then enhanced using the same methods as the radar branch. Each point $P_s$ in the enhanced pseudo-point cloud has the following attributes:
\begin{equation}
P_s = [x,y,z,x_c,y_c,z_c,x_p,y_p],
\end{equation}
where [\textit{x}, \textit{y}, \textit{z}] are the spatial coordinates of the pseudo point. 

After enhancing the pseudo-point cloud, point cloud data is processed on a pillar basis using random sampling or zero-padding to obtain the dense tensor $F_s$. Feature extraction is then performed following the \cref{equ_pseudo-point cloud}, resulting in a pseudo-image of size $(C_s, H, W)$, which serves as the BEV feature $F_{BEV}^S$ of the surface pseudo-point branch.
\begin{equation}
\label{equ_pseudo-point cloud}
F_{BEV}^{S} = Proj(MaxPool(LBR(F_s))).
\end{equation}

The BEV feature $F_{BEV}^S$, derived from the pseudo-point cloud generated through the surface fitting model, emphasizes foreground objects with dense spatial information, effectively complementing the sparse features of the 4D radar branch.

\subsection{Multi-layer Fusion and Detection Head}

\subsubsection{Multi-layer Fusion}
Following the BEV feature extraction from the image, radar, and surface pseudo-point branches, the multi-layer fusion module concatenates the features $F_{BEV}^I$, $ F_{BEV}^R$, $F_{BEV}^S$ and applies further feature extraction to produce the fused BEV feature $F_{BEV}$ with the shape $(C,H,W)$. The specific process of the features fusion is as follows:
\begin{equation}
F_{BEV} = CBR(Concat(F_{BEV}^I,F_{BEV}^R,F_{BEV}^S)).
\end{equation}

To enhance the detection of objects with diverse scales, we employ convolutional layers with multiple parameter settings to extract multi-scale feature maps, which provide richer representations for the subsequent detection head. 

\subsubsection{Detection Head}
We utilize the anchor-based detection head to predict 3D bounding boxes. The ground truth and predicted bounding boxes are defined as $[x,y,z,w,h,l,\theta]$, and the offset between the ground truth and predicted boxes can be defined as follows:
\begin{equation}
\begin{aligned}
    \Delta x &= \frac{x^{gt} - x^p}{\sqrt{(w^p)^2 + (l^p)^2}}, \quad
    \Delta y = \frac{y^{gt} - y^p}{\sqrt{(w^p)^2 + (l^p)^2}}, \quad \\
    \Delta z &= \frac{z^{gt} - z^p}{h^p}, \quad 
    \Delta w = \log \frac{w^{gt}}{w^p}, \quad 
    \Delta l = \log \frac{l^{gt}}{l^p}, \\
    \Delta h &= \log \frac{h^{gt}}{h^p}, \quad 
    \Delta \theta = \sin (\theta^{gt} - \theta^p),
\end{aligned}
\end{equation}
where subscripts $gt$ denotes the ground truth and $p$ denotes the predicted results. The localization loss $L_{loc}$ is formulated as follows:
\begin{equation}
L_{loc} = \frac{1}{N_{pos}} \sum_{i=1}^{N_{pos}} \sum_{t_i \in [x_i, y_i, z_i, w_i, l_i, h_i, \theta_i]} Smooth_{L1}(\Delta t_i),
\end{equation}
where $N_{pos}$ is the number of positive samples. The classification loss $L_{cls}$ is defined as:
\begin{equation}
L_{cls} = -\frac{1}{N_{pos}} \sum_{i=1}^{N_{pos}} \sum_{j=1}^{N_{cls}}\alpha_{i,j} (1 - p_{i,j})^\gamma \log p_{i,j},
\end{equation}
where $N_{cls}$ is the number of object categories, $p_{i,j}$ is the probability of an anchor belonging to the certain class, $\alpha_{i,j}$ is the weight factor, and $\gamma = 2$. The total loss is: 
\begin{equation}
L = \beta_1L_{loc} + \beta_2L_{cls} + \beta_3L_{dir},
\end{equation}
where $L_{dir}$ denotes the direction loss between the ground truth and predicted boxes, and $\beta_1$, $\beta_2$, and $\beta_3$ denotes the weight of each loss term.

\section{Experiment} \label{sec:experiment}

\subsection{Dataset Setup and Metric}
\textbf{Dataset:} In this study, we validate the performance of the proposed SFGFusion using two datasets: TJ4DRadSet \cite{tj4dradset} and VoD \cite{vod}. Both datasets are automotive datasets tailored for urban scenarios, containing various data from different sensors, including camera images and 4D imaging radar point clouds. Each dataset provides category labels, 2D bounding boxes, and 3D bounding boxes for every object. 

The TJ4DRadSet dataset encompasses a range of lighting conditions as well as various road types. This dataset contains 7,757 frames of synchronized and calibrated multi-sensor data in 44 sequences. For our experiments, we adopted the same label configuration as existing detection methods, which includes car, pedestrian, cyclist, and truck. The TJ4DRadSet dataset was split into a training set with 5,717 frames and a test set with 2,040 frames.

The main scenes in the VoD dataset include campus, suburb, and old-town areas in Delft, with a particular focus on scenarios involving vulnerable road users (VRUs), such as pedestrians and cyclists. The VoD dataset offers over 8,600 frames and more than 123,000 3D bounding box annotations, including the three categories used in our experiments: car, pedestrian, and cyclist. We followed the official data splits, dividing the dataset into a training set with 5,139 frames and a validation set with 1,296 frames. Additionally, the dataset provides an official testing code, which we used to obtain the final test results after training.

\textbf{Evaluation Metrics:}
We evaluated the performance of the SFGFusion using the specified evaluation parameters for each dataset. In the TJ4DRadSet dataset, we evaluated the detection results for objects within a 70m range from the radar source. The evaluation metrics include 3D AP and BEV AP for various categories of objects, as well as the mean Average Precision (mAP) for all categories. The Intersection over Union (IoU) thresholds for cars, pedestrians, cyclists, and trucks are set at 0.5, 0.25, 0.25, and 0.5, respectively.

In the VoD dataset, the official evaluation metrics include AP in the entire annotated area and AP in the region of interest. The region of interest is defined as a specific area in the camera coordinate system, specifically the driving corridor near the ego-vehicle, with the bounds $D_{RoI} = \{ (x,y,z) \mid -4m < x < 4m, z < 25m \}$. During AP calculation, the IoU thresholds for cars, pedestrians, and cyclists are the same as those in the TJ4DRadSet dataset.

\subsection{Implementation Details}

\textbf{Hyper-parameter Settings:}
The hyper-parameters in our experiments are aligned with the official specifications provided by the dataset. For the TJ4DRadSet dataset, the radar point cloud range is defined as 0m to 69.12m along the x-axis, -39.68m to 39.68m along the y-axis, and -4m to 2m along the z-axis. For the VoD dataset, the radar point cloud range is defined as 0m to 51.2m along the x-axis, -25.6m to 25.6m along the y-axis, and -3m to 2m along the z-axis. For both datasets, the pillar dimensions in the radar branch are set to 0.16 meters in both length and width, and the voxel dimensions during the voxelization process are set to 0.16m  $\times$ 0.16m $\times$ 6m for TJ4DRadSet and 0.16m $\times$ 0.16m $\times$ 5m for VoD. 

Moreover, we employ predefined anchor boxes in the detection head during the output of detection results. For the TJ4DRadSet dataset, the anchor box dimensions ($l$, $w$, $h$) for the ``Car", ``Pedestrian", ``Cyclist", and ``Truck" categories are set to (4.56m, 1.84m, 1.7m), (0.8m, 0.6m, 1.69m), (1.77m, 0.78m, 1.6m), and (10.76m, 2.66m, 3.47m), respectively. For the VoD dataset, the anchor box dimensions for the ``Car", ``Pedestrian", and ``Cyclist" categories are set to (3.9m, 1.6m, 1.56m), (0.8m, 0.6m, 1.73m), and (1.76m, 0.6m, 1.73m), respectively.

\textbf{Training Details:}
Our method is implemented based on the OpenPCDet \cite{openpcdet2020} framework and trained on four NVIDIA GeForce RTX 3090 GPUs. All branches of the model were optimized jointly during training. For the radar branch, we applied data augmentation strategies, including random world flipping along the x-axis, random world scaling (from 0.95 to 1.05 of the range), random world rotation (from -0.7854 to 0.7854 radians), and random world translation (Gaussian noise with a standard deviation of 0.5 on the x, y, and z axes). For the image branch, we applied corresponding processing to the images based on the data augmentation used in the radar branch. The instance masks used in the surface fitting model were generated using Mask2Former~\cite{mask2former}. During training, these masks were precomputed and loaded as part of the dataset to accelerate training. During inference, Mask2Former was executed online to ensure that the runtime performance evaluation reflects the actual system behavior. Although SFGFusion supports end-to-end training without pre-training the surface fitting network, we found that initialization influences detection performance. Therefore, we pre-train the surface fitting model and fix its parameters during SFGFusion training. For model optimization, we used the AdamW optimizer with an initial learning rate of 0.002 and trained for 40 epochs. 

\subsection{Experiment Results}

\begin{table*}[t!]
    \centering
    \caption{\textbf{Comparison of detectors' results on the TJ4DRadSet \cite{tj4dradset} dataset.} R denotes 4D imaging radar and C represents camera. Bold and underlined values denote the best and second-best performance, respectively. SFGFusion is implemented by augmenting the baseline with our proposed surface fitting model.}
    \fontsize{8pt}{8pt}\selectfont
    \setlength{\tabcolsep}{2pt}
    \label{Tab: TJ4DRadset Result}
    \begin{tabular}{cccccccccccc}
    \bottomrule[1pt]
    \multirow{2}{*}{\centering Method} & \multirow{2}{*}{\centering Modality} & \multicolumn{5}{c}{3D (\%)} & \multicolumn{5}{c}{BEV (\%)} \\
    \cline{3-12}
         & & Car & Ped & Cyc & Tru & mAP & Car & Ped & Cyc & Tru & mAP \\
    \hline
        ImVoxelNet \cite{imvoxelnet} & C & 22.55 & 13.73 & 9.67 & 13.87 & 14.96 & 26.10 & 14.21 & 10.99 & 17.18 & 17.12 \\
    \hline
        SECOND \cite{second} & R & 18.18 & 24.43 & 32.36 & 14.76 & 22.43 & 36.02 & 28.58 & 39.75 & 19.35 & 30.93 \\
        PointPillars \cite{pointpillars} & R & 21.26 & 28.33 & 52.47 & 11.18 & 28.31 & 38.34 & 32.26 & 56.11 & 18.19 & 36.23 \\
        Part-A$^2$ \cite{part-A} & R & 18.65 & 23.28 & 44.14 & 9.63 & 23.92 & 29.95 & 24.31 & 49.08 & 15.05 & 29.60 \\
        RPFA-Net \cite{RPFA-Net} & R & 26.89 & 27.36 & 50.95 & 14.46 & 29.91 & 42.89 & 29.81 & 57.09 & 25.98 & 38.94 \\
        CenterPoint \cite{centerpoint} & R & 22.03 & 25.02 & 53.32 & 15.92 & 29.07 & 33.03 & 27.87 & 58.74 & 25.09 & 36.18 \\
        VoxelNeXt \cite{centerpoint} & R & 13.27 & \textbf{33.54} & 52.59 & 8.32 & 26.93 & 23.17 & \textbf{35.83} & 57.11 & 12.12 & 32.06 \\
        PillarNeXt \cite{pillarnext} & R & 22.33 & 23.48 & 53.01 & 17.99 & 29.20 & 36.84 & 25.17 & 57.07 & 23.76 & 35.71 \\
        RadarPillarNet \cite{RCFusion} & R & 28.45 & 26.24 & 51.57 & 15.20 & 30.37 & 45.72 & 29.19 & 56.89 & 25.17 & 39.24 \\
        SMURF \cite{smurf} & R &  28.47 & 26.22 & 54.61 & 22.64 & 32.99 & 43.13  & 29.19 & 58.81 & 32.80 & 40.98 \\
    \hline
        MVX-Net \cite{mvx-net} & C+R & 22.28 & 19.57 & 50.70 & 11.21 & 25.94 & 37.46 & 22.70 & 54.69 & 18.07 & 33.23 \\
        PointAugmenting \cite{pointaugmenting} & C+R & 22.63 & 26.23 & 53.52 & 13.37 & 28.94 & \underline{43.42} & 29.65 & 59.21 & 23.88 & 39.04 \\
        Focals Conv \cite{focal} & C+R & 12.24 & \underline{31.80} & 54.01 & 6.66 & 26.18 & 22.52 & \underline{35.33} & \underline{59.37} & 10.30 & 31.88 \\
        RCFusion \cite{RCFusion} & C+R & \underline{29.72} & 27.17 & \underline{54.93} & 23.56 & 33.85 & 40.89 & 30.95 & 58.30 & 28.92 & 39.76 \\
        FUTR3D \cite{futr3d} & C+R & - & - & - & - & 32.42 & - & - & - & - & 37.51 \\
        BEVFusion \cite{bevfusion} & C+R & - & - & - & - & 32.71 & - & - & - & - & 41.12 \\
        LXL \cite{LXL} & C+R & - & - & - & - & \textbf{36.32} & - & - & - & - & \underline{41.20} \\
    \hline
        Baseline & C+R & 24.00 & 25.55 & 50.77 & \underline{24.62} & 31.23 & 34.37 & 27.31 & 54.64 & \underline{34.81} & 37.78 \\
        Baseline + SF & C+R & \textbf{33.05} & 27.01 &  \textbf{55.12} &  \textbf{27.10} & \underline{35.57} & \textbf{46.63} & 30.04 & \textbf{59.55} & \textbf{42.41} & \textbf{44.66} \\
         \textit{Improvement} & - & \textcolor{red}{\textit{+9.05}} & \textcolor{red}{\textit{+1.46}} & \textcolor{red}{\textit{+4.35}} & \textcolor{red}{\textit{+2.48}} & \textcolor{red}{\textit{+4.34}} & \textcolor{red}{\textit{+12.26}} & \textcolor{red}{\textit{+2.73}} & \textcolor{red}{\textit{+4.91}} & \textcolor{red}{\textit{+7.60}} & \textcolor{red}{\textit{+6.88}} \\
    \toprule[1pt]
    \end{tabular}    
\end{table*}

\textbf{Results on TJ4DRadSet:} To validate the performance of our proposed model, we conducted experiments using the TJ4DRadSet dataset\cite{tj4dradset} and compared its results with existing methods in \cref{Tab: TJ4DRadset Result}. SFGFusion, which augments the baseline with our proposed surface fitting model, demonstrates superior performance across almost all metrics compared to several leading detection methods. Specifically, it outperformed the second-best method, LXL, by 3.46\% in BEV mAP. These results demonstrate the effectiveness of the guidance by surface fitting in the SFGFusion. In both 3D AP and BEV AP, SFGFusion demonstrated higher performance to LXL in the car, cyclist, and truck categories, with notable increases of 3.33\% and 3.50\% in 3D AP and BEV AP for the car category, which has the most instances. Moreover, compared to the baseline, the integration of the surface fitting model into both the image feature view transformation and the surface pseudo-point branch significantly boosts the performance of SFGFusion, resulting in consistent accuracy improvements across all evaluation metrics. This demonstrates the potential of the surface fitting model as a general-purpose plug-in to enhance other fusion-based detection frameworks. 

When comparing algorithms based on different point cloud processing methods, it is evident that voxel-based algorithms, such as SECOND, perform slightly worse than pillar-based methods like PointPillars. This discrepancy arises from the imaging principles of 4D imaging radar, which generates far fewer points per frame compared to LiDAR. Voxel-based methods typically divide the 3D space into fine-grained regions, resulting in a large number of empty voxels, which limits effective feature extraction. Additionally, the vertical resolution of 4D imaging radar is lower than that of LiDAR. Pillar-based methods, which partition the 3D space into vertical columns of infinite height, do not rely on fine height distinctions, effectively mitigating the low vertical resolution of 4D imaging radar. This is why we opted for a pillar-based method in the radar branch and surface pseudo-point branch for feature extraction. 

The sparsity of 4D imaging radar point cloud affects not only the radar branch but also the performance of the surface fitting model. As evidenced by the results on the TJ4DRadSet dataset, the SFGFusion achieves notable performance advancements in the car and truck categories compared to other methods, with limited improvements in the pedestrian and cyclist categories. This is because the larger physical size of cars and trucks produces more radar reflection points, resulting in denser 4D radar point clouds that strengthen the depth estimation in the surface fitting model. In contrast, pedestrians and cyclists typically have only 2-3 radar points per object, and noise from radar signals along with sensor calibration errors hinder accurate supervision of surface parameter fitting, leading to potential depth estimation errors for these categories. For future work, we plan to incorporate dense LiDAR point clouds during training as a replacement for radar point clouds to improve supervision.

\begin{table}[t]
    \centering
    \caption{\textbf{The SFGFusion's evaluating results on different distances.}}
    \label{Tab: TJ4DRadset Distance Result}
    \fontsize{8pt}{8pt}\selectfont
    \setlength{\tabcolsep}{6pt}
    \begin{tabular}{ccccccc}
    \bottomrule[1pt]
    \multirow{2}{*}{\centering Label} & \multicolumn{3}{c}{3D (\%)} & \multicolumn{3}{c}{BEV (\%)} \\
    \cline{2-7}
         & 0-25m & 25-50m & 50-70m & 0-25m & 25-50m & 50-70m\\ 
    \hline
        Car & 37.79 & 29.89 & 27.05 & 47.14 & 46.21 & 37.49 \\
        Ped & 46.22 & 16.56 & 0.00 & 49.75 & 16.93 & 0.00 \\
        Cyc & 68.40 & 46.24 & 10.11 & 72.78 & 52.61 & 15.74 \\
        Tru & 36.83 & 32.68 & 16.47 & 47.75 & 46.42 & 31.66 \\
    \hline
        Avg & 47.31 & 31.34 & 13.41 & 54.36 & 40.54 & 21.22 \\
    \toprule[1pt]
    \end{tabular}
\end{table}

\begin{figure}[h]
\centering{\includegraphics[scale=0.33]{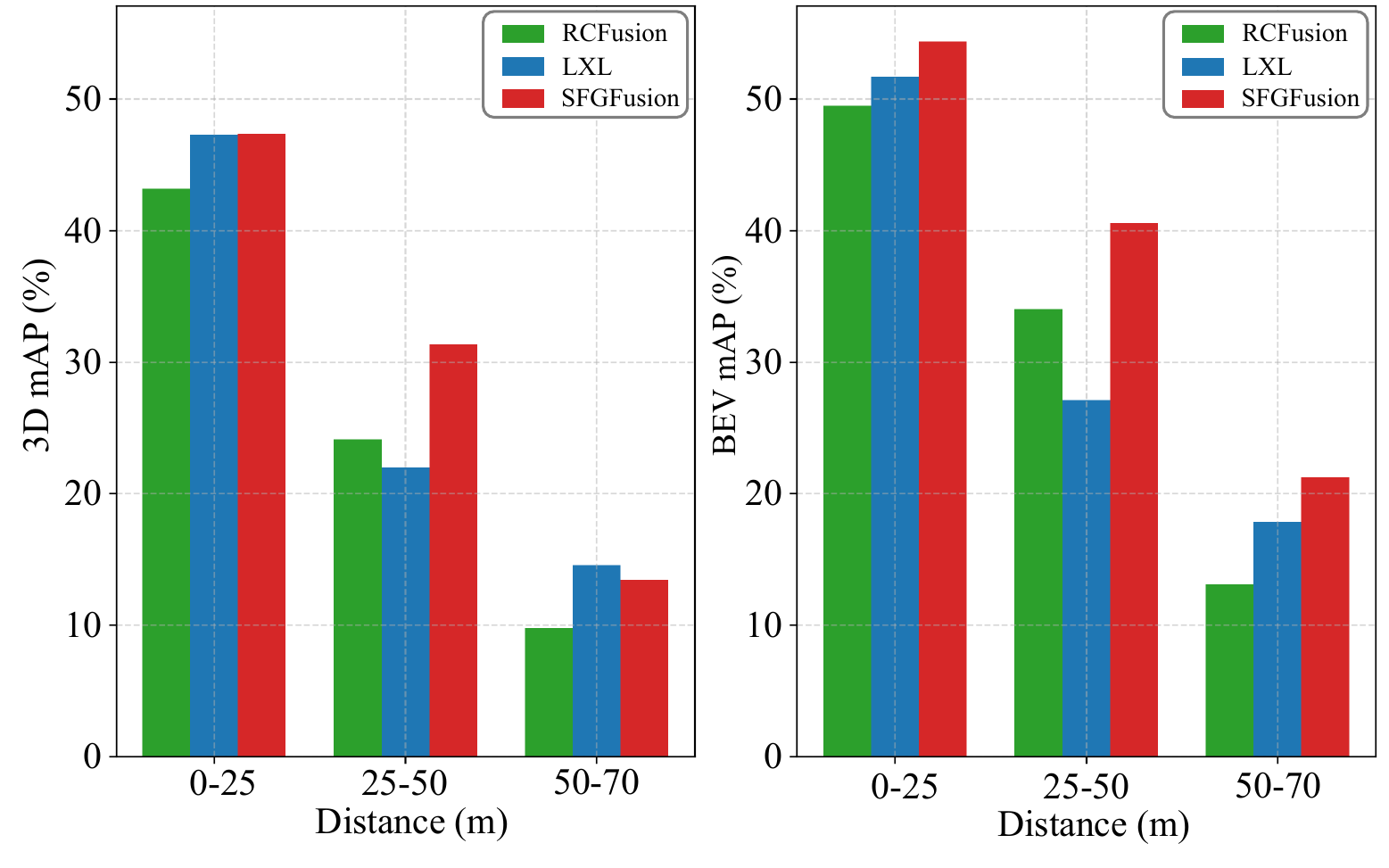}}
\caption{\textbf{Distance-based evaluation for camera-4D radar fusion detection methods.} Only methods with available distance-based evaluation results are included for fair and meaningful comparison.}
\label{Fig: TJ4DRadset Distance Result}
\end{figure}

We evaluated the detection accuracy of various object categories at different distances from the ego-vehicle, with results shown in \cref{Tab: TJ4DRadset Distance Result}. For short-range objects within 0-25 meters, our method achieved superior AP results across all categories. As the distance increased, detection accuracy declined across all object types. Notably, the precision for car and truck decreased less significantly, which can be explained by their larger physical size, allowing for more stable and denser visual and point cloud features even at greater distances. In contrast, pedestrian detection showed the most pronounced decline in accuracy at longer distances. This limitation can be attributed to two main factors. First, the pedestrian category is underrepresented in the dataset, particularly at long distances, which increases the difficulty of both training and prediction. Second, the radar point cloud is usually concentrated in areas with high reflectivity, while pedestrian surfaces are often smooth and lack metallic components, leading to sparse point cloud data at long distances, further complicating detection. Additionally, we also compared the mean detection accuracy across distances between SFGFusion and leading fusion-based methods. As shown in \cref{Fig: TJ4DRadset Distance Result}, SFGFusion consistently achieves higher 3D and BEV mAP values across almost all distance ranges. Particularly in BEV mAP, the surface fitting model enables precise depth estimation, yielding superior performance at all distances. For 3D mAP, SFGFusion is distinctly better than the LXL at the middle distance, and competitive at other distances.

\begin{figure*}[t]
\centering{\includegraphics[width=\textwidth, height=\textheight, keepaspectratio]{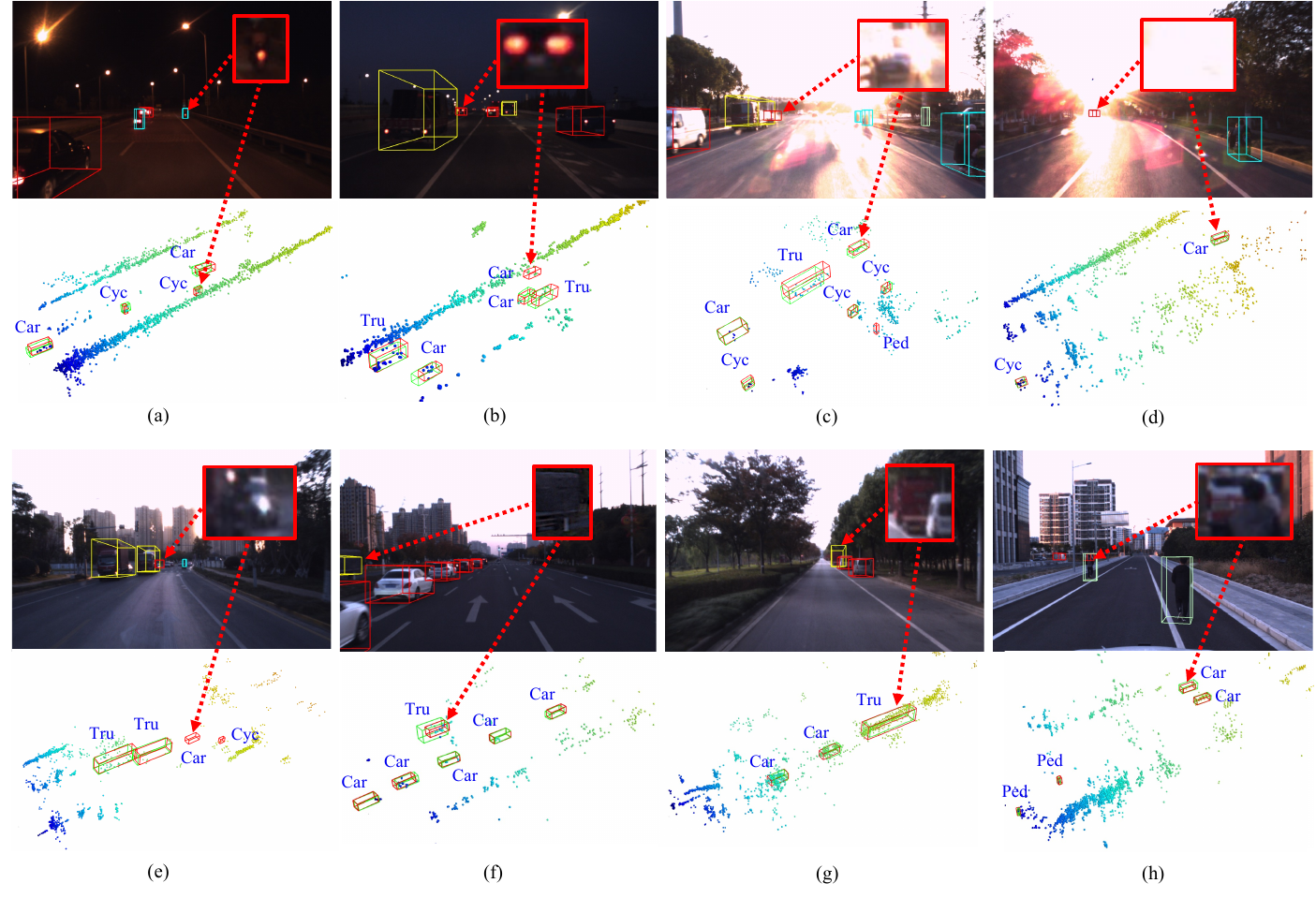}}
\caption{\textbf{The visualization of the SFGFusion's results on the TJ4DRadSet \cite{tj4dradset} test set.}
Our method performs robustly under challenging conditions such as darkness, strong lighting (a–d), and complex cases like long-range objects, occlusion, and truncation (e–h). Each scene includes visualizations of image and 3D point cloud. In the image, predicted object categories are distinguished using different colors. In the point cloud view, green and red bounding boxes denote ground truth and predictions, respectively.}
\label{Fig: TJ4DRad Result}
\end{figure*}

\cref{Fig: TJ4DRad Result} presents visualizations of the SFGFusion’s detection results on various scenes from the TJ4DRadSet test set. In challenging conditions like \cref{Fig: TJ4DRad Result} (a)-(d), where image quality is affected by darkness or intense lighting, certain objects are nearly undetectable using image-only methods. However, the algorithm consistently provides reliable detection results based on point cloud features. In addition, \cref{Fig: TJ4DRad Result} (e)-(h) demonstrates the model’s robustness in complex scenarios, including long-range detection, occlusion, and truncation. The extended sensing range of 4D imaging radar offers a significant advantage in capturing distant and partially visible objects.

\begin{table*}[t]
    \centering
    \caption{\textbf{Comparison of detectors' results on the VoD \cite{vod} dataset.} R denotes 4D imaging radar and C represents camera. Bold and underlined values denote the best and second-best performance, respectively. SFGFusion is implemented by augmenting the baseline with our proposed surface fitting model.}
    \label{Tab:VoD Result}
    \begin{threeparttable}
    \fontsize{8pt}{8pt}\selectfont
    \setlength{\tabcolsep}{2pt}
    \begin{tabular}{cc*{4}{>{\centering\arraybackslash}p{0.9cm}}*{4}{>{\centering\arraybackslash}p{0.8cm}}c}
    \bottomrule[1pt]
    \multirow{2}{*}{\centering Method} & \multirow{2}{*}{\centering Modality} & \multicolumn{4}{c}{AP in the Entire Annotated Area (\%)} & \multicolumn{4}{c}{AP in the Region of Interest (\%)} & \multirow{2}{*}{\centering FPS} \\
    \cline{3-10}
         & & Car & Ped & Cyc & mAP & Car & Ped & Cyc & mAP & \\
    \hline
        ImVoxelNet \cite{imvoxelnet} & C & 19.35 & 5.62 & 17.53 & 14.17 & 49.52 &  9.68 & 28.97 & 29.39 & 11.1 \\
    \hline
        PointPillars \cite{pointpillars} & R & 38.69 & 31.54 & 65.52 & 45.25 & 71.47 & 41.54 & 87.88 & 66.97 & 42.9 \\
         CenterPoint \cite{centerpoint} & R & 32.74 & 38.00 & 65.51 & 45.42 & 62.01 & 48.18 & 84.98 & 65.06 & 34.5 \\
         MVFAN \cite{mvfan} & R & 34.05 & 27.27 & 57.14 & 39.42 & 69.81 & 38.65 & 84.87 & 64.38 & 45.1\\
         PillarNeXt \cite{pillarnext} & R & 30.81 & 33.11 & 62.78 & 42.23 & 66.72 & 39.03 & 85.08 & 63.61 & 28.0 \\
         RadarPillarNet \cite{RCFusion} & R & 39.30 & 35.10 & 63.63 & 46.01 & 71.65 & 42.80 & 83.14 & 65.86 & \textit{N/A} \\
         SMIFormer \cite{smiformer} & R & 39.53 & 41.88 & 64.91 & 48.77 & 77.04 & 53.40 & 82.95 & 71.13 & 16.4 \\
         SMURF \cite{smurf} & R & 42.31 & 39.09 & 71.50 & 50.97 & 71.74 & 50.54 & 86.87 & 69.72 & 23.1 \\
         SCKD \cite{sckd} & R & 41.89 & 43.51 & 70.83 & 52.08 & 77.54 & 51.06 & 86.89 & 71.80 & 39.3 \\
    \hline
         BEVFusion \cite{bevfusion} & C+R & 42.02 & 38.98 & 67.54 & 49.51 & 72.23 & 48.67 & 85.57 & 69.02 & 8.4\\
         RCFusion \cite{RCFusion} & C+R & 41.70 & 38.95 & 68.31 & 49.65 & 71.87 & 47.50 & \underline{88.33} & 69.23 & 4.7\\
         FUTR3D \cite{futr3d} & C+R & \underline{46.01} & 35.11 & 65.98 & 46.84 & 78.66 & 43.10 & 86.19 & 69.32 & 7.3\\
         LXL \cite{LXL} & C+R & 42.33 &  \textbf{49.48} &  \textbf{77.12} &  \textbf{56.31} & 72.18 & \textbf{58.30} & 88.31 & 72.93 & 6.1\\
         RCBEVDet \cite{RCBevdet} & C+R & 40.63 & 38.86 & 70.48 & 49.99 & 72.48 & 49.89 & 87.01 & 69.80 & \textit{N/A}\\    
         ZFusion \cite{zfusion} & C+R & 43.89 & 39.48 & 70.46 & 51.28 & \textbf{79.51} & 52.95 & 86.37 & \underline{72.94} & 11.2\\
    \hline
        Baseline & C+R & 37.03 & 42.92 & 71.21 & 50.38 & 76.21 & 51.03 & 87.54 & 71.59 & 7.8\\
        Baseline + SF & C+R & \textbf{48.30} & \underline{43.60} & \underline{75.54} & \underline{55.76} &  \underline{79.10} & \underline{53.63} &  \textbf{88.60} &  \textbf{73.77} & 6.4\\
         \textit{Improvement} & - & \textcolor{red}{\textit{+11.27}} & \textcolor{red}{\textit{+0.68}} & \textcolor{red}{\textit{+4.33}} & \textcolor{red}{\textit{+5.38}} & \textcolor{red}{\textit{+2.89}} & \textcolor{red}{\textit{+2.60}} & \textcolor{red}{\textit{+1.06}} & \textcolor{red}{\textit{+2.18}} & -\\   
    \toprule[1pt]
    \end{tabular}
    \begin{tablenotes}    
    \footnotesize           
    \item[\textasteriskcentered]Region of Interest refers to the driving corridor near the ego-vehicle, defined in the camera coordinate as $D_{RoI} = \{ (x,y,z) \mid -4m < x < 4m, z < 25m \}$.
    \end{tablenotes}   
    \end{threeparttable}
\end{table*}

\textbf{Results on VoD:}
To further evaluate the performance of our proposed model, we conducted additional tests on the VoD dataset \cite{vod} as shown in \cref{Tab:VoD Result}. Our proposed SFGFusion significantly improves the baseline’s performance across all categories by incorporating the surface fitting model. Compared with other methods, SFGFusion consistently placed within the top two across all metrics, with significant improvements in the region of interest. This indicates that the surface fitting model allows the SFGFusion to more effectively focus on foreground information, accurately fitting the object surface parameters, and enhancing the view transformation and point cloud processing stages. 

We further assessed the inference speed of different approaches on the VoD dataset in \cref{Tab:VoD Result}. With the integration of image data, multimodal detection models face a significant increase in model size and computation, reducing the inference speed to below 10 fps. Despite this, our model has shown a notable improvement in performance, but there is no obvious decrease in the inference rate.

\begin{figure*}[t]
\centering{\includegraphics[width=\textwidth , height=\textheight, keepaspectratio]{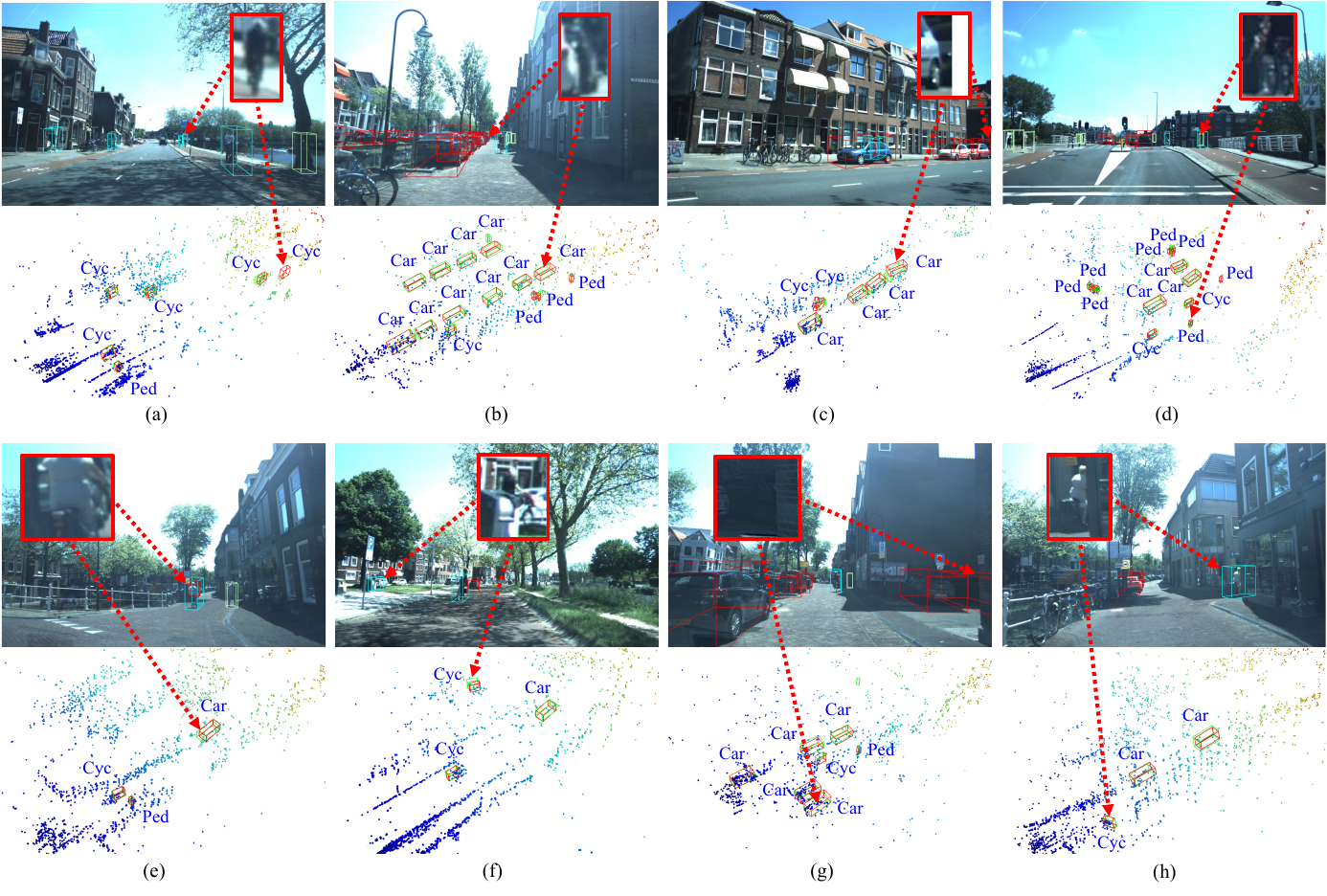}}
\caption{\textbf{The visualization of the SFGFusion's results on the VoD \cite{vod} test set.}
Our method accurately detects objects under occlusion, truncation, and complex backgrounds, and can identify objects absent from the ground truth (a) and extend the camera view (c). Each scene includes visualizations of image and 3D point cloud. In the image, predicted object categories are distinguished using different colors. In the point cloud view, green and red bounding boxes denote ground truth and predictions, respectively.}
\label{Fig: VoD Result}
\end{figure*}

\cref{Fig: VoD Result} presents SFGFusion visualizations across diverse scenes in the VoD dataset. The results show that the SFGFusion can accurately detect multiple objects despite occlusion, truncation, and complex backgrounds, and it demonstrates a strong capability for recognizing distant objects. Notably, in scene (c), the gray car extends beyond the camera view, yet the SFGFusion effectively employs radar data for successful recognition. Moreover, the SFGFusion can even correctly identify the unmarked distant object in scene (a) that is absent from the ground truth. Overall, our model efficiently integrates multi-modal information, leveraging the strengths of both camera and 4D imaging radar modalities to enhance detection performance.

\subsection{Ablation Experiments}

\begin{table*}[t]
    \centering
    \caption{\textbf{Ablation on key components with surface fitting.} Experiments are conducted on the TJ4DRadSet~\cite{tj4dradset} dataset.}
    \label{Tab: Ablation Surface Fitting Dual Branch}
    \begin{threeparttable}
    \fontsize{8pt}{10pt}\selectfont
    \setlength{\tabcolsep}{4pt}
    \begin{tabular}{ccccccccccccc}
    \bottomrule[1pt]    
    \multirow{2}{*}{\centering Baseline} & \multicolumn{2}{c}{Component} &  \multicolumn{5}{c}{3D (\%)} & \multicolumn{5}{c}{BEV (\%)} \\
    \cline{2-13}
         & SP & VT & Car & Ped & Cyc & Tru & mAP & Car & Ped & Cyc & Tru & mAP \\
    \hline
        \checkmark &  &  & 24.00 & 25.55 & 50.77 & 24.62 & 31.23 & 34.37 & 27.31 & 54.64 & 34.81 & 37.78 \\
        \checkmark & \checkmark &  & 27.12 & \underline{26.00} & \underline{54.77} & 25.43 & 33.33 & 37.93 & \underline{28.68} & \underline{56.51} & 35.33 & 39.61\\
        \checkmark &  & \checkmark & \underline{32.35} & 25.96 & 53.49 & \underline{25.92} & \underline{34.43} & \underline{45.35} & 28.47 & 55.87 & \underline{40.80} & \underline{42.62}\\
         \checkmark & \checkmark & \checkmark & \textbf{33.05} & \textbf{27.01} &  \textbf{55.12} &  \textbf{27.10} & \textbf{35.57} & \textbf{46.63} & \textbf{30.04} & \textbf{59.55} & \textbf{42.41} & \textbf{44.66}  \\
    \toprule[1pt]    
    \end{tabular}
    \begin{tablenotes}    
    \footnotesize              
    \item[*]SP: surface pseudo-point branch, VT: surface fitting information on image view transformation.
    \end{tablenotes}            
    \end{threeparttable}
\end{table*}

In this subsection, we conducted several ablation studies on the TJ4DRadSet and VoD datasets to validate the effectiveness of critical modules in SFGFusion. Specifically, we examined three aspects: the influence of surface fitting on subsequent image transformation and surface pseudo-point branch (\cref{Tab: Ablation Surface Fitting Dual Branch}); the effect of different surface parameter fitting methods within the surface fitting model (\cref{Tab: Ablation Surface Fitting}); and the contribution of multi-modal feature fusion (\cref{Tab: Modality Fusion}).

\textbf{Ablation on key components with surface fitting.} To evaluate the impact of the surface fitting model, we tested the effect of the dense depth information it generates on two key components: the image feature view transformation (VT) and the surface pseudo-point branch (SP). In addition to our method, which uses depth information to guide both components, we compared configurations where surface fitting guides only one or neither of them, as shown in \cref{Tab: Ablation Surface Fitting Dual Branch}. 

\begin{figure*}[t]
\centering{\includegraphics[width=\textwidth , height=\textheight, keepaspectratio]{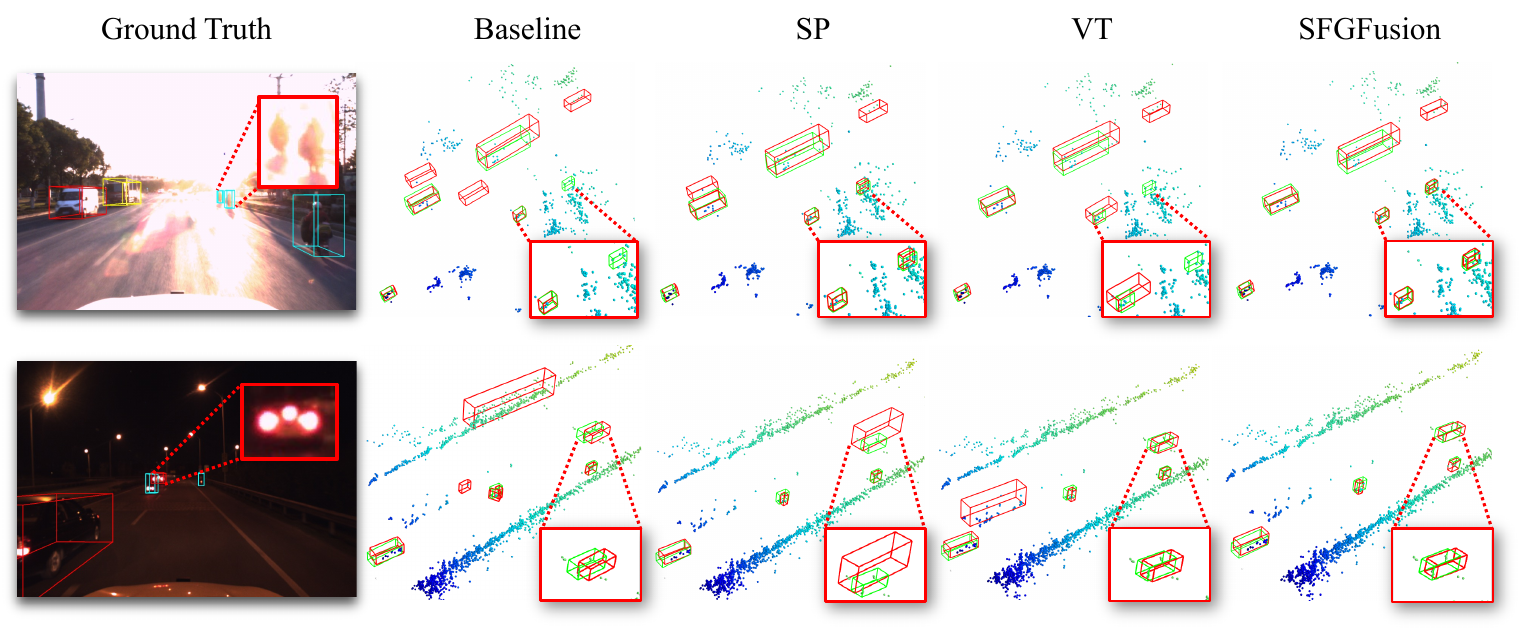}}
\caption{\textbf{The visualization of ablation studies on key components with surface fitting.} The SP component enhances the detection of small objects, while the VT component improves the localization accuracy of large objects. Combining SP and VT fully leverages the complementary strengths of surface fitting, resulting in the most accurate and robust detection performance. Each row corresponds to a different configuration: Baseline, SP, VT, and SFGFusion. In the point cloud view, green and red bounding boxes denote ground truth and predictions, respectively.}
\label{Fig: Ablation vis}
\end{figure*}

The results indicate that the SP component effectively improves detection accuracy, particularly for small objects such as pedestrians and cyclists. This improvement arises because such objects typically contain fewer point clouds, and the pseudo-point cloud generated from surface fitting effectively compensates for the sparsity of 4D imaging radar data, thereby enhancing detection. The VT component also yields notable accuracy gains, especially for large objects such as cars and trucks. Since these objects naturally contain more point clouds, the performance boost from the SP component is relatively less pronounced, while the VT component can more precisely project image feature to the correct spatial location under the guidance of surface fitting, offering greater detection benefits. Overall, the SP and VT components complement each other, effectively exploiting the surface fitting information to deliver substantial improvements over the baseline model. Furthermore, we visualize the experimental results in \cref{Fig: Ablation vis}. These findings demonstrate that guiding both SP and VT with surface fitting maximizes the complementary advantages of multi-modal fusion, yielding significant overall performance gains in 3D object detection.

\textbf{Ablation on surface fitting methods.} To validate the effectiveness of our surface fitting method, we conducted experiments from three perspectives: whether to apply surface fitting, the choice of surface shape, and the method for surface parameter fitting. For the surface fitting application comparison, we removed the surface fitting model entirely, preventing it from participating in the image feature view transformation and surface pseudo-point branch. For the surface shape comparison, we performed surface fitting using plane and quadratic surface. For the parameter fitting method, we evaluated three strategies: (1) constructing a fitting plane based on the average depth value of all point clouds within an instance; (2) fitting surface parameters using the least squares method with point cloud depth; (3) employing our deep learning approach that integrates both image and point cloud information for parameter prediction.

\begin{table*}[t]
    \centering
    \caption{\textbf{Ablation on surface fitting methods.} Experiments are conducted on the TJ4DRadSet \cite{tj4dradset} dataset.}
    \label{Tab: Ablation Surface Fitting}
    \begin{threeparttable}
    \fontsize{8pt}{10pt}\selectfont
    \setlength{\tabcolsep}{1.3pt}
    \begin{tabular}{ccccccccccccc}
    \bottomrule[1pt]
    \multirow{2}{*}{\makecell{Surface\\Fitting}} & \multirow{2}{*}{\makecell{Surface\\Shape}} & \multirow{2}{*}{\makecell{Fitting\\Method}} & \multicolumn{5}{c}{3D (\%)} & \multicolumn{5}{c}{BEV (\%)} \\
    \cline{4-13}
         & & & Car & Ped & Cyc & Tru & mAP & Car & Ped & Cyc & Tru & mAP \\
    \hline
        \(\times\) & - & - & 24.00 & 25.55 & 50.77 & 24.62 & 31.23 & 34.37 & 27.31 & 54.64 & 34.81 & 37.78 \\
        \checkmark& Plane & Average Depth & 26.22 & 25.38 & 54.60 & 26.10 & 33.08 & 43.55 & 27.50 & 58.98 & 37.73 & 41.94 \\
        \checkmark& Plane &  Least Squares & 31.51 & 25.96 & 53.07 & \textbf{27.68} & 34.56 & 44.81 & 28.13 & 56.57 & 39.27 & 42.19 \\
        \checkmark& Plane &  Deep Learning & \underline{31.66} & 26.44 & 55.06 & 26.21 & \underline{34.84} & \underline{45.61} & 29.91 & \underline{59.20} & 39.33 & \underline{43.51}  \\
        \checkmark & Quadratic Surface & Least Squares & 31.02 & \underline{26.64} & \textbf{55.46} & 25.49 & 34.65 & 44.05 & \textbf{30.61} & 59.15 & \underline{39.44} & 43.31 \\
         \checkmark& Quadratic Surface & Deep Learning & \textbf{33.05} & \textbf{27.01} &  \underline{55.12} & \underline{27.10} & \textbf{35.57} & \textbf{46.63} & \underline{30.04} & \textbf{59.55} & \textbf{42.41} & \textbf{44.66}  \\
    \toprule[1pt]
    \end{tabular}          
    \end{threeparttable}
\end{table*}

Based on the results in \cref{Tab: Ablation Surface Fitting}, our proposed deep learning-based quadratic surface fitting method effectively leverages image semantics and point cloud geometric information, significantly improving the model performance. In the comparison of fitting methods, under the same surface shape conditions, the least squares method, which only uses point cloud depth without incorporating image semantics, produced less accurate surface parameter estimates than the deep learning method. The average depth method, which assigns the mean depth of all points within an instance to every pixel, was the least effective because it simplifies the surface to a plane with uniform depth. In terms of surface shape, the plane representation, due to its limited parameters, struggles to accurately capture the true surface shape of objects, resulting in poorer performance. In contrast, the quadratic surface, with higher degrees of freedom, can more precisely model the curved morphology of objects, thereby achieving better detection performance under the same fitting method conditions. Moreover, completely removing the surface fitting model caused a significant drop in performance, clearly demonstrating the critical role of this module within the overall detection framework.

\textbf{Ablation on multi-modal feature fusion.} To validate the effectiveness of multi-modal fusion, we evaluated the performance of SFGFusion using the camera-only and radar-only modalities separately. As shown in \cref{Tab: Modality Fusion}, the algorithm using only the camera modality significantly lags behind other methods. This deficiency stems primarily from the absence of depth information in monocular images, which leads to inaccuracies in estimating the 3D spatial positions of objects. Moreover, our camera branch is mainly designed for BEV feature extraction and lacks specific adaptations tailored for the 3D detection task. Conversely, the radar-only algorithm demonstrates better performance, but its efficacy is still limited by the sparse nature of the radar point cloud. When fusing both image and radar modalities, detection performance improves substantially across all metrics. This indicates that our proposed SFGFusion method effectively facilitates multi-modal interaction, leveraging the dense semantic and color information from image alongside the radar’s precise spatial and velocity measurements to significantly enhance 3D object detection accuracy.

\begin{table}[t]
    \centering
    \caption{\textbf{Ablation on multi-modal feature fusion.} R denotes 4D imaging radar and C represents camera.}
    \label{Tab: Modality Fusion}
    \fontsize{9pt}{10pt}\selectfont
    \setlength{\tabcolsep}{10pt}
    \begin{tabular}{ccccc}
    \bottomrule[1pt]
    \multirow{2}{*}{\centering Modality} & \multicolumn{2}{c}{TJ4D mAP (\%)} & \multicolumn{2}{c}{VoD mAP (\%)} \\
    \cline{2-5}
         & 3D & BEV & EAA & RoI\\ 
    \hline
         C & 7.71 & 9.18 & 8.22 & 15.10 \\
         R & 29.57 & 35.83 & 48.44 & 67.03 \\
         C+R & \textbf{35.57} & \textbf{44.66} & \textbf{55.76} & \textbf{73.77}  \\ 
        \textit{Improvement} & \textcolor{red}{\textit{+6.00}} & \textcolor{red}{\textit{+8.83}} & \textcolor{red}{\textit{+7.32}} & \textcolor{red}{\textit{+6.74}}  \\

    \toprule[1pt]
    \end{tabular}
\end{table}

\section{Conclusion} \label{sec:conclusion}

In this paper, we introduced SFGFusion, a novel network for 3D object detection that fuses camera and 4D imaging radar data. Experimental results on the TJ4DRadSet and VoD datasets confirm that the SFGFusion surpasses the performance of existing methods. This performance advantage is attributed to the surface fitting model, which effectively leverages image and sparse radar points to predict object surface parameters. These parameters enable accurate prediction of object depth, with the resulting depth information guiding image feature transformation and pseudo-point cloud extraction in subsequent components, leading to significant improvements in detection performance.

As one of the pioneering efforts to leverage 4D imaging radar for multi-modal detection, SFGFusion makes a significant contribution to the field, laying a solid foundation for future advancements in multi-modal fusion techniques. Given the modularity and deployment flexibility of the surface fitting model, future work will further explore its performance potential and optimize it into a plug-and-play software module to evaluate its applicability and effectiveness within other fusion algorithms. Additionally, we plan to explore fusion methods for multi-modal data after feature extraction, enhancing efficient interaction and integration of different modalities. 

\section{Acknowledgments}
This work was supported by the National Natural Science Foundation of China [grant number 62172043] and the Key Project of Science and Technology Development Fund of China Coal Research Institute CCTEG [grant number 2024ZDI-12].

\bibliographystyle{elsarticle-num-names}


\begin{thebibliography}{49}
\expandafter\ifx\csname natexlab\endcsname\relax\def\natexlab#1{#1}\fi
\providecommand{\url}[1]{\texttt{#1}}
\providecommand{\href}[2]{#2}
\providecommand{\path}[1]{#1}
\providecommand{\DOIprefix}{doi:}
\providecommand{\ArXivprefix}{arXiv:}
\providecommand{\URLprefix}{URL: }
\providecommand{\Pubmedprefix}{pmid:}
\providecommand{\doi}[1]{\href{http://dx.doi.org/#1}{\path{#1}}}
\providecommand{\Pubmed}[1]{\href{pmid:#1}{\path{#1}}}
\providecommand{\bibinfo}[2]{#2}
\ifx\xfnm\relax \def\xfnm[#1]{\unskip,\space#1}\fi
\bibitem[{Mao et~al.(2023)Mao, Shi, Wang, and Li}]{survey_autonomous_driving}
\bibinfo{author}{J.~Mao}, \bibinfo{author}{S.~Shi}, \bibinfo{author}{X.~Wang}, \bibinfo{author}{H.~Li},
\newblock \bibinfo{title}{{3D} object detection for autonomous driving: A comprehensive survey},
\newblock \bibinfo{journal}{Int. J. Comput. Vis.} \bibinfo{volume}{131} (\bibinfo{year}{2023}) \bibinfo{pages}{1909--1963}.
\bibitem[{Fan et~al.(2024)Fan, Wang, Chang, Li, Wang, and Cao}]{survey_4d_radar}
\bibinfo{author}{L.~Fan}, \bibinfo{author}{J.~Wang}, \bibinfo{author}{Y.~Chang}, \bibinfo{author}{Y.~Li}, \bibinfo{author}{Y.~Wang}, \bibinfo{author}{D.~Cao},
\newblock \bibinfo{title}{{4D} mmwave radar for autonomous driving perception: A comprehensive survey},
\newblock \bibinfo{journal}{{IEEE} Trans. Intell. Veh.} \bibinfo{volume}{9} (\bibinfo{year}{2024}) \bibinfo{pages}{4606--4620}.
\bibitem[{Zheng et~al.(2023)Zheng, Li, Tan, Yang, Chen, Huang, Bai, Zhu, and Ma}]{RCFusion}
\bibinfo{author}{L.~Zheng}, \bibinfo{author}{S.~Li}, \bibinfo{author}{B.~Tan}, \bibinfo{author}{L.~Yang}, \bibinfo{author}{S.~Chen}, \bibinfo{author}{L.~Huang}, \bibinfo{author}{J.~Bai}, \bibinfo{author}{X.~Zhu}, \bibinfo{author}{Z.~Ma},
\newblock \bibinfo{title}{{RCFusion}: Fusing 4-{D} radar and camera with bird’s-eye view features for 3-{D} object detection},
\newblock \bibinfo{journal}{{IEEE} Trans. Instrum. Meas.} \bibinfo{volume}{72} (\bibinfo{year}{2023}) \bibinfo{pages}{1--14}.
\bibitem[{Xiong et~al.(2024)Xiong, Liu, Huang, Han, Xia, and Zhu}]{LXL}
\bibinfo{author}{W.~Xiong}, \bibinfo{author}{J.~Liu}, \bibinfo{author}{T.~Huang}, \bibinfo{author}{Q.-L. Han}, \bibinfo{author}{Y.~Xia}, \bibinfo{author}{B.~Zhu},
\newblock \bibinfo{title}{{LXL}: Lidar excluded lean {3D} object detection with {4D} imaging radar and camera fusion},
\newblock \bibinfo{journal}{{IEEE} Trans. Intell. Veh.}  (\bibinfo{year}{2024}) \bibinfo{pages}{3142--3142}.
\bibitem[{Lin et~al.(2024)Lin, Liu, Xia, Wang, Wang, Qi, Dong, Dong, Zhang, and Zhu}]{RCBevdet}
\bibinfo{author}{Z.~Lin}, \bibinfo{author}{Z.~Liu}, \bibinfo{author}{Z.~Xia}, \bibinfo{author}{X.~Wang}, \bibinfo{author}{Y.~Wang}, \bibinfo{author}{S.~Qi}, \bibinfo{author}{Y.~Dong}, \bibinfo{author}{N.~Dong}, \bibinfo{author}{L.~Zhang}, \bibinfo{author}{C.~Zhu},
\newblock \bibinfo{title}{{RCBEVDet}: Radar-camera fusion in bird's eye view for {3D} object detection},
\newblock in: \bibinfo{booktitle}{Proc. IEEE/CVF Conf. Comput. Vis. Pattern Recognit. (CVPR)}, \bibinfo{year}{2024}, pp. \bibinfo{pages}{14928--14937}.
\bibitem[{Qian et~al.(2022)Qian, Lai, and Li}]{survey_3D_detection}
\bibinfo{author}{R.~Qian}, \bibinfo{author}{X.~Lai}, \bibinfo{author}{X.~Li},
\newblock \bibinfo{title}{{3D} object detection for autonomous driving: A survey},
\newblock \bibinfo{journal}{Pattern Recognit.} \bibinfo{volume}{130} (\bibinfo{year}{2022}) \bibinfo{pages}{108796}.
\bibitem[{Liu et~al.(2023)Liu, Tang, Amini, Yang, Mao, Rus, and Han}]{bevfusion}
\bibinfo{author}{Z.~Liu}, \bibinfo{author}{H.~Tang}, \bibinfo{author}{A.~Amini}, \bibinfo{author}{X.~Yang}, \bibinfo{author}{H.~Mao}, \bibinfo{author}{D.~L. Rus}, \bibinfo{author}{S.~Han},
\newblock \bibinfo{title}{{BEVFusion}: Multi-task multi-sensor fusion with unified bird's-eye view representation},
\newblock in: \bibinfo{booktitle}{Proc. IEEE Int. Conf. Rob. Autom. (ICRA)}, \bibinfo{year}{2023}, pp. \bibinfo{pages}{2774--2781}.
\bibitem[{Chen et~al.(2022)Chen, Li, Zhang, Sun, and Jia}]{focal}
\bibinfo{author}{Y.~Chen}, \bibinfo{author}{Y.~Li}, \bibinfo{author}{X.~Zhang}, \bibinfo{author}{J.~Sun}, \bibinfo{author}{J.~Jia},
\newblock \bibinfo{title}{Focal sparse convolutional networks for 3d object detection},
\newblock in: \bibinfo{booktitle}{Proc. IEEE/CVF Conf. Comput. Vis. Pattern Recognit. (CVPR)}, \bibinfo{year}{2022}, pp. \bibinfo{pages}{5428--5437}.
\bibitem[{Qi et~al.(2017)Qi, Su, Mo, and Guibas}]{pointnet}
\bibinfo{author}{C.~R. Qi}, \bibinfo{author}{H.~Su}, \bibinfo{author}{K.~Mo}, \bibinfo{author}{L.~J. Guibas},
\newblock \bibinfo{title}{{PointNet}: Deep learning on point sets for {3D} classification and segmentation},
\newblock in: \bibinfo{booktitle}{Proc. IEEE/CVF Conf. Comput. Vis. Pattern Recognit. (CVPR)}, \bibinfo{year}{2017}, pp. \bibinfo{pages}{652--660}.
\bibitem[{Lang et~al.(2019)Lang, Vora, Caesar, Zhou, Yang, and Beijbom}]{pointpillars}
\bibinfo{author}{A.~H. Lang}, \bibinfo{author}{S.~Vora}, \bibinfo{author}{H.~Caesar}, \bibinfo{author}{L.~Zhou}, \bibinfo{author}{J.~Yang}, \bibinfo{author}{O.~Beijbom},
\newblock \bibinfo{title}{{PointPillars}: Fast encoders for object detection from point clouds},
\newblock in: \bibinfo{booktitle}{Proc. IEEE/CVF Conf. Comput. Vis. Pattern Recognit. (CVPR)}, \bibinfo{year}{2019}, pp. \bibinfo{pages}{12697--12705}.
\bibitem[{Zhou and Tuzel(2018)}]{Voxelnet}
\bibinfo{author}{Y.~Zhou}, \bibinfo{author}{O.~Tuzel},
\newblock \bibinfo{title}{{VoxelNet}: End-to-end learning for point cloud based {3D} object detection},
\newblock in: \bibinfo{booktitle}{Proc. IEEE/CVF Conf. Comput. Vis. Pattern Recognit. (CVPR)}, \bibinfo{year}{2018}, pp. \bibinfo{pages}{4490--4499}.
\bibitem[{Zheng et~al.(2022)Zheng, Ma, Zhu, Tan, Li, Long, Sun, Chen, Zhang, Wan et~al.}]{tj4dradset}
\bibinfo{author}{L.~Zheng}, \bibinfo{author}{Z.~Ma}, \bibinfo{author}{X.~Zhu}, \bibinfo{author}{B.~Tan}, \bibinfo{author}{S.~Li}, \bibinfo{author}{K.~Long}, \bibinfo{author}{W.~Sun}, \bibinfo{author}{S.~Chen}, \bibinfo{author}{L.~Zhang}, \bibinfo{author}{M.~Wan}, et~al.,
\newblock \bibinfo{title}{{TJ4DRadSet}: A {4D} radar dataset for autonomous driving},
\newblock in: \bibinfo{booktitle}{IEEE Conf. Intell. Transport. Syst. Proc. (ITSC)}, \bibinfo{year}{2022}, pp. \bibinfo{pages}{493--498}.
\bibitem[{Palffy et~al.(2022)Palffy, Pool, Baratam, Kooij, and Gavrila}]{vod}
\bibinfo{author}{A.~Palffy}, \bibinfo{author}{E.~Pool}, \bibinfo{author}{S.~Baratam}, \bibinfo{author}{J.~F. Kooij}, \bibinfo{author}{D.~M. Gavrila},
\newblock \bibinfo{title}{Multi-class road user detection with 3+1{D} radar in the view-of-delft dataset},
\newblock \bibinfo{journal}{{IEEE} Robot. Autom. Lett.} \bibinfo{volume}{7} (\bibinfo{year}{2022}) \bibinfo{pages}{4961--4968}.
\bibitem[{Rukhovich et~al.(2022)Rukhovich, Vorontsova, and Konushin}]{imvoxelnet}
\bibinfo{author}{D.~Rukhovich}, \bibinfo{author}{A.~Vorontsova}, \bibinfo{author}{A.~Konushin},
\newblock \bibinfo{title}{{ImVoxelNet}: Image to voxels projection for monocular and multi-view general-purpose {3D} object detection},
\newblock in: \bibinfo{booktitle}{Proc. IEEE/CVF Winter Conf. Appl. Comput. Vis. Workshops (WACVW)}, \bibinfo{year}{2022}, pp. \bibinfo{pages}{2397--2406}.
\bibitem[{Linghu and Ling(2025)}]{adaptive}
\bibinfo{author}{J.~Linghu}, \bibinfo{author}{Q.~Ling},
\newblock \bibinfo{title}{Adaptive depth position encoding for sparse query-based {3D} object detection from multi-camera images},
\newblock \bibinfo{journal}{Pattern Recognit.}  (\bibinfo{year}{2025}) \bibinfo{pages}{111747}.
\bibitem[{Li et~al.(2022)Li, Wang, Li, Xie, Sima, Lu, Qiao, and Dai}]{bevformer}
\bibinfo{author}{Z.~Li}, \bibinfo{author}{W.~Wang}, \bibinfo{author}{H.~Li}, \bibinfo{author}{E.~Xie}, \bibinfo{author}{C.~Sima}, \bibinfo{author}{T.~Lu}, \bibinfo{author}{Y.~Qiao}, \bibinfo{author}{J.~Dai},
\newblock \bibinfo{title}{{BEVFormer}: Learning bird’s-eye-view representation from multi-camera images via spatiotemporal transformers},
\newblock in: \bibinfo{booktitle}{Proc. Eur. Conf. Comput. Vis. (ECCV)}, \bibinfo{year}{2022}, pp. \bibinfo{pages}{1--18}.
\bibitem[{Yang et~al.(2023)Yang, Chen, Tian, Tao, Zhu, Zhang, Huang, Li, Qiao, Lu, Zhou, and Dai}]{bevformerv2}
\bibinfo{author}{C.~Yang}, \bibinfo{author}{Y.~Chen}, \bibinfo{author}{H.~Tian}, \bibinfo{author}{C.~Tao}, \bibinfo{author}{X.~Zhu}, \bibinfo{author}{Z.~Zhang}, \bibinfo{author}{G.~Huang}, \bibinfo{author}{H.~Li}, \bibinfo{author}{Y.~Qiao}, \bibinfo{author}{L.~Lu}, \bibinfo{author}{J.~Zhou}, \bibinfo{author}{J.~Dai},
\newblock \bibinfo{title}{{BEVFormer} v2: Adapting modern image backbones to bird's-eye-view recognition via perspective supervision},
\newblock in: \bibinfo{booktitle}{Proc. IEEE/CVF Conf. Comput. Vis. Pattern Recognit. (CVPR)}, \bibinfo{year}{2023}, pp. \bibinfo{pages}{17830--17839}.
\bibitem[{Wang et~al.(2019)Wang, Chao, Garg, Hariharan, Campbell, and Weinberger}]{pseudo-lidar}
\bibinfo{author}{Y.~Wang}, \bibinfo{author}{W.-L. Chao}, \bibinfo{author}{D.~Garg}, \bibinfo{author}{B.~Hariharan}, \bibinfo{author}{M.~Campbell}, \bibinfo{author}{K.~Q. Weinberger},
\newblock \bibinfo{title}{Pseudo-lidar from visual depth estimation: Bridging the gap in 3d object detection for autonomous driving},
\newblock in: \bibinfo{booktitle}{Proc. IEEE/CVF Conf. Comput. Vis. Pattern Recognit. (CVPR)}, \bibinfo{year}{2019}, pp. \bibinfo{pages}{8445--8453}.
\bibitem[{Tao et~al.(2023)Tao, Cao, Wang, Zhang, and Gao}]{pseudo-mono}
\bibinfo{author}{C.~Tao}, \bibinfo{author}{J.~Cao}, \bibinfo{author}{C.~Wang}, \bibinfo{author}{Z.~Zhang}, \bibinfo{author}{Z.~Gao},
\newblock \bibinfo{title}{Pseudo-{Mono} for monocular {3D} object detection in autonomous driving},
\newblock \bibinfo{journal}{{IEEE} Trans. Circuits Syst. Video Technol.} \bibinfo{volume}{33} (\bibinfo{year}{2023}) \bibinfo{pages}{3962--3975}.
\bibitem[{Li et~al.(2023)Li, Luo, and Yang}]{pillarnext}
\bibinfo{author}{J.~Li}, \bibinfo{author}{C.~Luo}, \bibinfo{author}{X.~Yang},
\newblock \bibinfo{title}{{PillarNeXt}: Rethinking network designs for {3D} object detection in lidar point clouds},
\newblock in: \bibinfo{booktitle}{Proc. IEEE/CVF Conf. Comput. Vis. Pattern Recognit. (CVPR)}, \bibinfo{year}{2023}, pp. \bibinfo{pages}{17567--17576}.
\bibitem[{Ma et~al.(2024)Ma, Huang, Qian, Kang, Liu, Zhang, and Hong}]{lgnet}
\bibinfo{author}{J.~Ma}, \bibinfo{author}{Y.~Huang}, \bibinfo{author}{C.~Qian}, \bibinfo{author}{J.~Kang}, \bibinfo{author}{J.~Liu}, \bibinfo{author}{H.~Zhang}, \bibinfo{author}{W.~Hong},
\newblock \bibinfo{title}{{LGNet}: Local and global point dependency network for {3D} object detection},
\newblock \bibinfo{journal}{Pattern Recognit.} \bibinfo{volume}{154} (\bibinfo{year}{2024}) \bibinfo{pages}{110585}.
\bibitem[{Lu et~al.(2024)Lu, Sun, Yang, Song, Jiang, and Liu}]{hrnet}
\bibinfo{author}{B.~Lu}, \bibinfo{author}{Y.~Sun}, \bibinfo{author}{Z.~Yang}, \bibinfo{author}{R.~Song}, \bibinfo{author}{H.~Jiang}, \bibinfo{author}{Y.~Liu},
\newblock \bibinfo{title}{{HRNet}: 3d object detection network for point cloud with hierarchical refinement},
\newblock \bibinfo{journal}{Pattern Recognit.} \bibinfo{volume}{149} (\bibinfo{year}{2024}) \bibinfo{pages}{110254}.
\bibitem[{Liu et~al.(2024)Liu, Zhao, Xiong, Huang, Han, and Zhu}]{smurf}
\bibinfo{author}{J.~Liu}, \bibinfo{author}{Q.~Zhao}, \bibinfo{author}{W.~Xiong}, \bibinfo{author}{T.~Huang}, \bibinfo{author}{Q.-L. Han}, \bibinfo{author}{B.~Zhu},
\newblock \bibinfo{title}{{SMURF}: Spatial multi-representation fusion for {3D} object detection with {4D} imaging radar},
\newblock \bibinfo{journal}{{IEEE} Trans. Intell. Veh.} \bibinfo{volume}{9} (\bibinfo{year}{2024}) \bibinfo{pages}{799--812}.
\bibitem[{Yan et~al.(2018)Yan, Mao, and Li}]{second}
\bibinfo{author}{Y.~Yan}, \bibinfo{author}{Y.~Mao}, \bibinfo{author}{B.~Li},
\newblock \bibinfo{title}{{SECOND}: Sparsely embedded convolutional detection},
\newblock \bibinfo{journal}{Sensors} \bibinfo{volume}{18} (\bibinfo{year}{2018}) \bibinfo{pages}{3337}.
\bibitem[{Yin et~al.(2021)Yin, Zhou, and Krahenbuhl}]{centerpoint}
\bibinfo{author}{T.~Yin}, \bibinfo{author}{X.~Zhou}, \bibinfo{author}{P.~Krahenbuhl},
\newblock \bibinfo{title}{Center-based {3D} object detection and tracking},
\newblock in: \bibinfo{booktitle}{Proc. IEEE/CVF Conf. Comput. Vis. Pattern Recognit. (CVPR)}, \bibinfo{year}{2021}, pp. \bibinfo{pages}{11784--11793}.
\bibitem[{Wang et~al.(2021)Wang, Chen, Deng, and Zhang}]{3d-centernet}
\bibinfo{author}{Q.~Wang}, \bibinfo{author}{J.~Chen}, \bibinfo{author}{J.~Deng}, \bibinfo{author}{X.~Zhang},
\newblock \bibinfo{title}{{3D-CenterNet}: {3D} object detection network for point clouds with center estimation priority},
\newblock \bibinfo{journal}{Pattern Recognit.} \bibinfo{volume}{115} (\bibinfo{year}{2021}) \bibinfo{pages}{107884}.
\bibitem[{Li et~al.(2022)Li, Yao, Quan, Xie, and Yang}]{spatial}
\bibinfo{author}{Z.~Li}, \bibinfo{author}{Y.~Yao}, \bibinfo{author}{Z.~Quan}, \bibinfo{author}{J.~Xie}, \bibinfo{author}{W.~Yang},
\newblock \bibinfo{title}{Spatial information enhancement network for {3D} object detection from point cloud},
\newblock \bibinfo{journal}{Pattern Recognit.} \bibinfo{volume}{128} (\bibinfo{year}{2022}) \bibinfo{pages}{108684}.
\bibitem[{Bang et~al.(2024)Bang, Choi, Kim, Kum, and Choi}]{radardistill}
\bibinfo{author}{G.~Bang}, \bibinfo{author}{K.~Choi}, \bibinfo{author}{J.~Kim}, \bibinfo{author}{D.~Kum}, \bibinfo{author}{J.~W. Choi},
\newblock \bibinfo{title}{{RadarDistill}: Boosting radar-based object detection performance via knowledge distillation from lidar features},
\newblock in: \bibinfo{booktitle}{Proc. IEEE/CVF Conf. Comput. Vis. Pattern Recognit. (CVPR)}, \bibinfo{year}{2024}, pp. \bibinfo{pages}{15491--15500}.
\bibitem[{Xu et~al.(2021)Xu, Zhang, Wang, Hu, Li, Pan, Li, and Deng}]{RPFA-Net}
\bibinfo{author}{B.~Xu}, \bibinfo{author}{X.~Zhang}, \bibinfo{author}{L.~Wang}, \bibinfo{author}{X.~Hu}, \bibinfo{author}{Z.~Li}, \bibinfo{author}{S.~Pan}, \bibinfo{author}{J.~Li}, \bibinfo{author}{Y.~Deng},
\newblock \bibinfo{title}{{RPFA-Net}: A {4D} radar pillar feature attention network for {3D} object detection},
\newblock in: \bibinfo{booktitle}{IEEE Conf. Intell. Transport. Syst. Proc. (ITSC)}, \bibinfo{year}{2021}, pp. \bibinfo{pages}{3061--3066}.
\bibitem[{Tan et~al.(2022)Tan, Ma, Zhu, Li, Zheng, Chen, Huang, and Bai}]{RadarMFNet}
\bibinfo{author}{B.~Tan}, \bibinfo{author}{Z.~Ma}, \bibinfo{author}{X.~Zhu}, \bibinfo{author}{S.~Li}, \bibinfo{author}{L.~Zheng}, \bibinfo{author}{S.~Chen}, \bibinfo{author}{L.~Huang}, \bibinfo{author}{J.~Bai},
\newblock \bibinfo{title}{3-{D} object detection for multiframe 4-{D} automotive millimeter-wave radar point cloud},
\newblock \bibinfo{journal}{{IEEE} Sensors J.} \bibinfo{volume}{23} (\bibinfo{year}{2022}) \bibinfo{pages}{11125--11138}.
\bibitem[{Wang et~al.(2021)Wang, Ma, Zhu, and Yang}]{pointaugmenting}
\bibinfo{author}{C.~Wang}, \bibinfo{author}{C.~Ma}, \bibinfo{author}{M.~Zhu}, \bibinfo{author}{X.~Yang},
\newblock \bibinfo{title}{{PointAugmenting}: Cross-modal augmentation for {3D} object detection},
\newblock in: \bibinfo{booktitle}{Proc. IEEE/CVF Conf. Comput. Vis. Pattern Recognit. (CVPR)}, \bibinfo{year}{2021}, pp. \bibinfo{pages}{11794--11803}.
\bibitem[{Li et~al.(2025)Li, Si, Liang, An, Tian, Zhou, and Wang}]{multimodal}
\bibinfo{author}{J.~Li}, \bibinfo{author}{G.~Si}, \bibinfo{author}{X.~Liang}, \bibinfo{author}{Z.~An}, \bibinfo{author}{P.~Tian}, \bibinfo{author}{F.~Zhou}, \bibinfo{author}{X.~Wang},
\newblock \bibinfo{title}{Multimodal fusion via voting network for {3D} object detection in indoors},
\newblock \bibinfo{journal}{Pattern Recognit.} \bibinfo{volume}{164} (\bibinfo{year}{2025}) \bibinfo{pages}{111501}.
\bibitem[{Chen et~al.(2017)Chen, Ma, Wan, Li, and Xia}]{mv3d}
\bibinfo{author}{X.~Chen}, \bibinfo{author}{H.~Ma}, \bibinfo{author}{J.~Wan}, \bibinfo{author}{B.~Li}, \bibinfo{author}{T.~Xia},
\newblock \bibinfo{title}{Multi-view {3D} object detection network for autonomous driving},
\newblock in: \bibinfo{booktitle}{Proc. IEEE/CVF Conf. Comput. Vis. Pattern Recognit. (CVPR)}, \bibinfo{year}{2017}, pp. \bibinfo{pages}{1907--1915}.
\bibitem[{Sindagi et~al.(2019)Sindagi, Zhou, and Tuzel}]{mvx-net}
\bibinfo{author}{V.~A. Sindagi}, \bibinfo{author}{Y.~Zhou}, \bibinfo{author}{O.~Tuzel},
\newblock \bibinfo{title}{{MVX-Net}: Multimodal voxelnet for {3D} object detection},
\newblock in: \bibinfo{booktitle}{Proc. IEEE Int. Conf. Rob. Autom. (ICRA)}, \bibinfo{organization}{IEEE}, \bibinfo{year}{2019}, pp. \bibinfo{pages}{7276--7282}.
\bibitem[{Wu et~al.(2022)Wu, Peng, Yang, Xie, Huang, Deng, Liu, and Cai}]{SFD}
\bibinfo{author}{X.~Wu}, \bibinfo{author}{L.~Peng}, \bibinfo{author}{H.~Yang}, \bibinfo{author}{L.~Xie}, \bibinfo{author}{C.~Huang}, \bibinfo{author}{C.~Deng}, \bibinfo{author}{H.~Liu}, \bibinfo{author}{D.~Cai},
\newblock \bibinfo{title}{Sparse fuse dense: Towards high quality {3D} detection with depth completion},
\newblock in: \bibinfo{booktitle}{Proc. IEEE/CVF Conf. Comput. Vis. Pattern Recognit. (CVPR)}, \bibinfo{year}{2022}, pp. \bibinfo{pages}{5418--5427}.
\bibitem[{Wu et~al.(2023)Wu, Wen, Shi, Li, and Wang}]{virtual-sparse}
\bibinfo{author}{H.~Wu}, \bibinfo{author}{C.~Wen}, \bibinfo{author}{S.~Shi}, \bibinfo{author}{X.~Li}, \bibinfo{author}{C.~Wang},
\newblock \bibinfo{title}{Virtual sparse convolution for multimodal {3D} object detection},
\newblock in: \bibinfo{booktitle}{Proc. IEEE/CVF Conf. Comput. Vis. Pattern Recognit. (CVPR)}, \bibinfo{year}{2023}, pp. \bibinfo{pages}{21653--21662}.
\bibitem[{He et~al.(2017)He, Gkioxari, Doll{\'a}r, and Girshick}]{mask-rcnn}
\bibinfo{author}{K.~He}, \bibinfo{author}{G.~Gkioxari}, \bibinfo{author}{P.~Doll{\'a}r}, \bibinfo{author}{R.~Girshick},
\newblock \bibinfo{title}{Mask {R-CNN}},
\newblock in: \bibinfo{booktitle}{Proc. IEEE/CVF Int. Conf. Comput. Vis. (ICCV)}, \bibinfo{year}{2017}, pp. \bibinfo{pages}{2961--2969}.
\bibitem[{Liu et~al.(2021)Liu, Lin, Cao, Hu, Wei, Zhang, Lin, and Guo}]{Swin-T}
\bibinfo{author}{Z.~Liu}, \bibinfo{author}{Y.~Lin}, \bibinfo{author}{Y.~Cao}, \bibinfo{author}{H.~Hu}, \bibinfo{author}{Y.~Wei}, \bibinfo{author}{Z.~Zhang}, \bibinfo{author}{S.~Lin}, \bibinfo{author}{B.~Guo},
\newblock \bibinfo{title}{Swin transformer: Hierarchical vision transformer using shifted windows},
\newblock in: \bibinfo{booktitle}{Proc. IEEE/CVF Int. Conf. Comput. Vis. (ICCV)}, \bibinfo{year}{2021}, pp. \bibinfo{pages}{10012--10022}.
\bibitem[{Lin et~al.(2017)Lin, Doll{\'a}r, Girshick, He, Hariharan, and Belongie}]{FPN}
\bibinfo{author}{T.-Y. Lin}, \bibinfo{author}{P.~Doll{\'a}r}, \bibinfo{author}{R.~Girshick}, \bibinfo{author}{K.~He}, \bibinfo{author}{B.~Hariharan}, \bibinfo{author}{S.~Belongie},
\newblock \bibinfo{title}{Feature pyramid networks for object detection},
\newblock in: \bibinfo{booktitle}{Proc. IEEE/CVF Conf. Comput. Vis. Pattern Recognit. (CVPR)}, \bibinfo{year}{2017}, pp. \bibinfo{pages}{2117--2125}.
\bibitem[{Philion and Fidler(2020)}]{lss}
\bibinfo{author}{J.~Philion}, \bibinfo{author}{S.~Fidler},
\newblock \bibinfo{title}{Lift, {Splat}, {Shoot}: Encoding images from arbitrary camera rigs by implicitly unprojecting to {3D}},
\newblock in: \bibinfo{booktitle}{Proc. Eur. Conf. Comput. Vis. (ECCV)}, \bibinfo{year}{2020}, pp. \bibinfo{pages}{194--210}.
\bibitem[{Huang et~al.(2021)Huang, Huang, Zhu, Ye, and Du}]{bevdet}
\bibinfo{author}{J.~Huang}, \bibinfo{author}{G.~Huang}, \bibinfo{author}{Z.~Zhu}, \bibinfo{author}{Y.~Ye}, \bibinfo{author}{D.~Du}, \bibinfo{title}{{BEVDet}: High-performance multi-camera {3D} object detection in bird-eye-view}, \bibinfo{year}{2021}. \bibinfo{note}{ArXiv preprint arXiv:2112.11790}.
\bibitem[{Team(2020)}]{openpcdet2020}
\bibinfo{author}{O.~D. Team}, \bibinfo{title}{{OpenPCDet}: An open-source toolbox for {3D} object detection from point clouds}, \bibinfo{howpublished}{\url{https://github.com/open-mmlab/OpenPCDet}}, \bibinfo{year}{2020}.
\bibitem[{Cheng et~al.(2022)Cheng, Misra, Schwing, Kirillov, and Girdhar}]{mask2former}
\bibinfo{author}{B.~Cheng}, \bibinfo{author}{I.~Misra}, \bibinfo{author}{A.~G. Schwing}, \bibinfo{author}{A.~Kirillov}, \bibinfo{author}{R.~Girdhar},
\newblock \bibinfo{title}{Masked-attention mask transformer for universal image segmentation},
\newblock in: \bibinfo{booktitle}{Proc. IEEE/CVF Conf. Comput. Vis. Pattern Recognit. (CVPR)}, \bibinfo{year}{2022}, pp. \bibinfo{pages}{1290--1299}.
\bibitem[{Shi et~al.(2020)Shi, Wang, Shi, Wang, and Li}]{part-A}
\bibinfo{author}{S.~Shi}, \bibinfo{author}{Z.~Wang}, \bibinfo{author}{J.~Shi}, \bibinfo{author}{X.~Wang}, \bibinfo{author}{H.~Li},
\newblock \bibinfo{title}{From points to parts: {3D} object detection from point cloud with part-aware and part-aggregation network},
\newblock \bibinfo{journal}{{IEEE} Trans. Pattern Anal. Mach. Intell.} \bibinfo{volume}{43} (\bibinfo{year}{2020}) \bibinfo{pages}{2647--2664}.
\bibitem[{Chen et~al.(2023)Chen, Zhang, Wang, Wang, and Zhao}]{futr3d}
\bibinfo{author}{X.~Chen}, \bibinfo{author}{T.~Zhang}, \bibinfo{author}{Y.~Wang}, \bibinfo{author}{Y.~Wang}, \bibinfo{author}{H.~Zhao},
\newblock \bibinfo{title}{{FUTR3D}: A unified sensor fusion framework for {3D} detection},
\newblock in: \bibinfo{booktitle}{Proc. IEEE/CVF Conf. Comput. Vis. Pattern Recognit. (CVPR)}, \bibinfo{year}{2023}, pp. \bibinfo{pages}{172--181}.
\bibitem[{Yan and Wang(2023)}]{mvfan}
\bibinfo{author}{Q.~Yan}, \bibinfo{author}{Y.~Wang},
\newblock \bibinfo{title}{Mvfan: Multi-view feature assisted network for {4D} radar object detection},
\newblock in: \bibinfo{booktitle}{Int. Conf. Neural Inf. Process. (ICONIP)}, \bibinfo{organization}{Springer}, \bibinfo{year}{2023}, pp. \bibinfo{pages}{493--511}.
\bibitem[{Shi et~al.(2023)Shi, Zhu, Zhang, Chen, Yu, and Zhu}]{smiformer}
\bibinfo{author}{W.~Shi}, \bibinfo{author}{Z.~Zhu}, \bibinfo{author}{K.~Zhang}, \bibinfo{author}{H.~Chen}, \bibinfo{author}{Z.~Yu}, \bibinfo{author}{Y.~Zhu},
\newblock \bibinfo{title}{{SMIFormer}: Learning spatial feature representation for {3D} object detection from {4D} imaging radar via multi-view interactive transformers},
\newblock \bibinfo{journal}{Sensors} \bibinfo{volume}{23} (\bibinfo{year}{2023}) \bibinfo{pages}{9429}.
\bibitem[{Xu et~al.(2025)Xu, Xiang, Zhang, Zhong, Zhao, Dang, Xu, Pu, and Liu}]{sckd}
\bibinfo{author}{R.~Xu}, \bibinfo{author}{Z.~Xiang}, \bibinfo{author}{C.~Zhang}, \bibinfo{author}{H.~Zhong}, \bibinfo{author}{X.~Zhao}, \bibinfo{author}{R.~Dang}, \bibinfo{author}{P.~Xu}, \bibinfo{author}{T.~Pu}, \bibinfo{author}{E.~Liu},
\newblock \bibinfo{title}{{SCKD}: Semi-supervised cross-modality knowledge distillation for {4D} radar object detection},
\newblock in: \bibinfo{booktitle}{Proc. AAAI Conf. Artif. Intell. (AAAI)}, \bibinfo{year}{2025}, pp. \bibinfo{pages}{8933--8941}.
\bibitem[{Yang et~al.(2025)Yang, Zhan, Qiao, Gong, Yang, Wang, and Lu}]{zfusion}
\bibinfo{author}{S.~Yang}, \bibinfo{author}{T.~Zhan}, \bibinfo{author}{S.~Qiao}, \bibinfo{author}{J.~Gong}, \bibinfo{author}{Q.~Yang}, \bibinfo{author}{J.~Wang}, \bibinfo{author}{Y.~Lu},
\newblock \bibinfo{title}{Zfusion: An effective fuser of camera and {4D} radar for {3D} object perception in autonomous driving},
\newblock in: \bibinfo{booktitle}{Proc. IEEE/CVF Conf. Comput. Vis. Pattern Recognit. Workshops (CVPRW)}, \bibinfo{year}{2025}, pp. \bibinfo{pages}{3768--3777}.

\end{thebibliography}
\end{document}